\documentclass[10pt,twocolumn,letterpaper]{article}

\usepackage[pagenumbers]{cvpr}              

\usepackage{amsmath}
\usepackage{amssymb}
\usepackage{graphicx}
\usepackage{siunitx}
\usepackage{makecell}
\usepackage{booktabs}
\usepackage{multirow}
\usepackage{bm}

\DeclareMathOperator{\normalize}{normalize} 
\DeclareMathOperator{\score}{score}

\definecolor{cvprblue}{rgb}{0.21,0.49,0.74}
\usepackage[pagebackref,breaklinks,colorlinks,allcolors=cvprblue]{hyperref}

\usepackage{amsmath}         
\usepackage{amssymb}         
\usepackage{algorithm}       
\usepackage[noend]{algpseudocode} 
\usepackage[capitalize,nameinlink]{cleveref} 
\usepackage{booktabs}
\usepackage{tabularx}
\usepackage{multicol}
\usepackage{framed}

\def\confName{CVPR}
\def\confYear{2026}

\title{\textbf{Taming Preference Mode Collapse via Directional Decoupling Alignment \\ in Diffusion Reinforcement Learning}}

\author{
    Chubin Chen$^{1,*, \S}$ \quad 
    Sujie Hu$^{1,*}$ \quad 
    Jiashu Zhu$^{2,*}$ \quad 
    Meiqi Wu$^2$ \quad
    Jintao Chen$^2$ \\
    Yanxun Li$^2$ \quad
    Nisha Huang$^1$ \quad
    Chengyu Fang$^1$ \quad
    Jiahong Wu$^{2,\ddagger}$ \quad 
    Xiangxiang Chu$^2$ \quad 
    Xiu Li$^{1,\dagger}$%
    \vspace{2pt} \\
    {\normalsize 
    $^1$Tsinghua University \quad 
    $^2$AMAP, Alibaba Group \quad
    } \\    
    {\small $^*$Equal contribution \quad $^\dagger$Corresponding author \quad $^\ddagger$Project lead \quad $^\S$Work done during the internship at AMAP} \\
}

\newcommand{\ours}{Directional Decoupling Alignment\xspace}
\newcommand{\ourshort}{D$^2$-Align\xspace}
\newcommand{\bench}{DivGenBench\xspace}
\newcommand{\supp}{Supp.\xspace}

\newcommand{\vx}{\bm{x}}
\newcommand{\vv}{\bm{v}}
\newcommand{\epsilonv}{\bm{\epsilon}}

\newcommand{\gN}{\mathcal{N}}
\newcommand{\E}{\mathbb{E}}

\newcommand{\rmI}{\mathbf{I}}

\newcolumntype{T}{>{\centering\arraybackslash}p{2.5cm}} 

\begin{document}

\twocolumn[{
    \renewcommand\twocolumn[1][]{#1}%
    \maketitle
    \vspace{-30pt}
    \begin{center}
        \centering
        \includegraphics[width=1.0\textwidth]{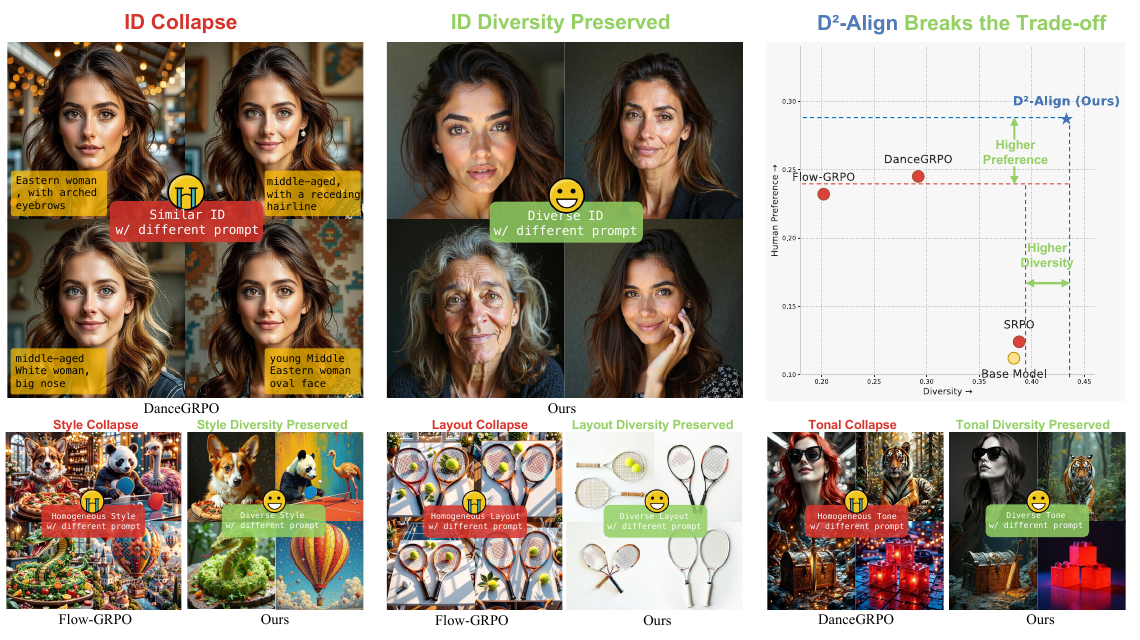}
            \captionof{figure}{\textbf{\ourshort breaks the trade-off between human preference and generative diversity, mitigating Preference Mode Collapse (PMC).}
            The top-right plot shows that while baselines struggle with a trade-off—either achieving low diversity or low preference—\ourshort, achieves a state of both higher diversity and higher human preference. The qualitative examples below illustrate this phenomenon. For the same set of varied prompts, baseline methods exhibit severe PMC, generating homogeneous outputs for identity, style, layout, and tone. \ourshort successfully preserves diversity, generating distinct and high-quality images that align with each individual prompt. See \supp for detail prompts.}
    \label{fig: teaser}
    \end{center}
}]


\begin{abstract} 
Recent studies have demonstrated significant progress in aligning text-to-image diffusion models with human preference via Reinforcement Learning from Human Feedback. 
However, while existing methods achieve high scores on automated reward metrics, they often lead to \textbf{Preference Mode Collapse (PMC)}—a specific form of reward hacking where models converge on narrow, high-scoring outputs (e.g., images with monolithic styles or pervasive overexposure), severely degrading generative diversity.
In this work, we introduce and quantify this phenomenon, proposing \textbf{\bench}, a novel benchmark designed to measure the extent of PMC. We posit that this collapse is driven by over-optimization along the reward model's inherent biases.
Building on this analysis, we propose \textbf{Directional Decoupling Alignment (D$^2$-Align)}, a novel framework that mitigates PMC by directionally correcting the reward signal.
Specifically, our method first learns a directional correction within the reward model's embedding space while keeping the model frozen.
This correction is then applied to the reward signal during the optimization process, preventing the model from collapsing into specific modes and thereby maintaining diversity.
Our comprehensive evaluation, combining qualitative analysis with quantitative metrics for both quality and diversity, reveals that \ourshort achieves superior alignment with human preference.
\end{abstract}

\section{Introduction}
\label{sec:intro}

Recent advances in generative models have enabled the generation of high-fidelity and creative visual content~\cite{diffusion,ddpm,cos, rfsolver,flow1, flow2, instantswap, genreaware,Huang_2025_ICCV, chen2025storyctrl, lei2025, wu2025imagerysearch,mao2025omni,xu2025scalar,chu2025usp}.
To better align model outputs with human preference, a pivotal approach has been the adoption of Reinforcement Learning from Human Feedback (RLHF)~\cite{rlhf1,rlhf2}.
A successful alignment process must consider not only image quality and fidelity~\cite{diffusion2, chen2025s2guidancestochasticselfguidance}, but also generative diversity~\cite{diversity1,diversity2}.
Key motivations for ensuring \textit{generative diversity} include its foundational role in creative content generation~\cite{diverse-need1}, data augmentation~\cite{diverse-need2}, and the enhancement of downstream task performance~\cite{diverse-need3}.
In practice, training T2I models to maximize a predefined reward often leads to reward hacking, where models learn to achieve high scores even for low-quality outputs~\cite{refl,flowgrpo}.
While existing methods have made progress in mitigating this outright quality degradation~\cite{refl,draft,flowgrpo}, they tend to steer the model toward a narrow high-score template~\cite{cfg++,apg}.
As shown in Fig.~\ref{fig: teaser},
this convergence produces highly homogeneous images characterized by a monolithic style, recurring visual features, or pervasive overexposure, severely degrading generative diversity.
We term this phenomenon \textbf{Preference Mode Collapse (PMC)}.
This issue stems from two primary challenges: (1) Current approaches prioritize image quality, largely overlooking the crucial aspect of output diversity. Furthermore, there is a lack of standardized quantitative metrics for its evaluation; and (2) the methods used to counteract reward hacking are often empirical and hyperparameter-sensitive, serving only to temper its magnitude rather than fundamentally altering the model's optimization direction.

Regarding the first challenge, existing methods for tackling reward hacking predominantly focus on image quality~\cite{refl,flowgrpo,dancegrpo}. While they strive to maintain quality as the reward increases, the optimization process inadvertently drives the model into what we term PMC.
This problem is compounded because, unlike fidelity \cite{pickscore, refl}, generative diversity is inherently difficult to quantify. 
To address this, we re-evaluate the reward hacking problem through the dual lenses of quality and diversity. Furthermore, we propose \textbf{\bench}, a novel benchmark designed to quantitatively measure the extent of this collapse, thereby offering a standardized means to assess generative diversity.

Regarding the second challenge, existing approaches to mitigate reward hacking often tackle the problem from distinct perspectives. For instance, Flow-GRPO~\cite{flowgrpo} employs KL divergence to prevent over-optimization, but this method requires extensive manual tuning of its coefficient and often leads to a time-consuming training process. Another approach, DanceGRPO~\cite{dancegrpo}, involves ensembling multiple reward models; however, its sensitivity to the ensemble weights and thresholds can result in inconsistent performance and training instability. 
Despite these different strategies, a shared limitation is that they primarily modulate the reward's magnitude, leaving the problematic optimization direction unaddressed. 
We hypothesize that the inherent biases of reward models are a primary cause of this phenomenon. 
To address this, we propose \textbf{\ours (D$^2$-Align)},  a novel optimization framework that directionally corrects the reward signal to mitigate PMC.
Specifically, our method first learns a directional correction within the reward model's embedding space while keeping the generator frozen.
During the subsequent optimization phase, applying the learned directional vector to correct the reward signal prevents the model from over-optimizing into specific modes, which in turn mitigates the collapse in diversity and guides the model toward a more genuine alignment with human preference.

In summary, our main contributions are as follows:
\begin{itemize}
    \item We introduce and quantify \textit{Preference Mode Collapse (PMC)}—a specific form of reward hacking where existing methods cause a severe loss of diversity when aligning with human preference. To this end, we propose \bench, a novel benchmark designed to serve as the primary tool for its measurement.
    
    \item We propose \textit{\ours (\ourshort)}, a novel framework designed to mitigate PMC by directionally correcting the reward signal, which maintains high quality and preserves diversity throughout optimization to achieve a more genuine alignment with human preference.
    
    \item Extensive quantitative and qualitative experiments, further corroborated by human evaluations, demonstrate that our method significantly outperforms existing approaches in terms of both generation quality and diversity.
\end{itemize}


\section{Related Work}
\label{sec:related}
\begin{figure*}[!t]
  \centering
    \includegraphics[width=1.0\textwidth]{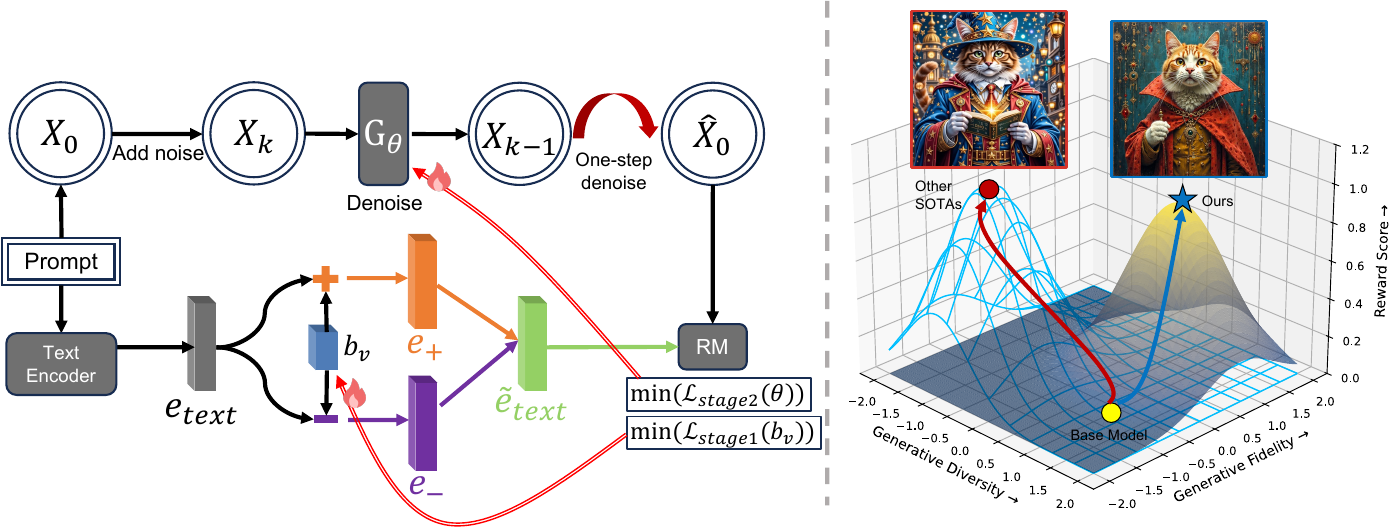}
    \caption{
        \textbf{Overview of \ourshort}. 
        \textbf{(Left)} Our framework first learns a directional vector ($\boldsymbol{b}_v$) to correct the reward signal while keeping the generator frozen (Stage 1). 
        It then uses this learned direction to guide the optimization of the generator, steering it away from mode collapse (Stage 2). 
        \textbf{(Right)} This visualization shows that while other methods converge to a narrow peak (low diversity), \ourshort finds a superior optimum that balances both quality and generative diversity.
    }

    \vspace{-10pt}
    \label{fig: pipeline}
\end{figure*}
\subsection{RL for Image Generation}
The application of reinforcement learning (RL) to T2I generation has seen significant advancements. 
Early works in this area can be broadly categorized into two primary directions. 
The first~\cite{rl,rl2-dpok,policy1,diversity2,policy2,policy3} integrated RL into diffusion models by optimizing the score function through policy gradient methods. An alternative approach~\cite{rlhf1,refl,differentiable1,differentiable2} focuses on direct fine-tuning with differentiable rewards. 
A key advantage of these methods is their use of analytical gradients rather than policy gradients, which allows for more efficient alignment.
More recently, methods inspired by the principles of GRPO~\cite{flowgrpo,dancegrpo,prefgrpo,branchgrpo,tempflowgrpo,mixgrpo, chu2025gpg,li2025adacurl} have elevated the performance to a new state-of-the-art, achieving substantially higher scores on given reward metrics. 
However, this success is accompanied by a significant challenge: reward hacking, where models over-optimize for superficial reward cues, producing outputs that misalign with true human preference.
\vspace{-10pt}

\subsection{Human Preference Alignment}
The inherent discrepancy between reward signals and true human preference poses a significant challenge, often leading to a degradation in both generation quality and diversity. 
From a quality perspective, DanceGRPO~\cite{dancegrpo} observed that combining rewards like HPS-v2.1~\cite{hps} and CLIP~\cite{hessel2021clipscore} can mitigate their respective failure modes. 
Pref-GRPO~\cite{prefgrpo} attributes this issue to the model's exploitation of subtle imperfections within the reward function, while MixGRPO~\cite{mixgrpo} alleviates it through hybrid inference at test time.
From a diversity standpoint, while Flow-GRPO's~\cite{flowgrpo} proposed KL-regularization term can counteract the decline in sample diversity, its practical application is hampered by several limitations. The method not only relies on a coefficient that must be manually tuned via trial and error, but also introduces significant training overhead and is highly sensitive to this choice. 
Furthermore, a quantitative evaluation of diversity is notably absent from their framework.
Crucially, prior works tend to address quality and diversity in isolation and do not provide a quantitative evaluation of diversity in the final, optimized model. 
In this work, we introduce a solution that holistically addresses both challenges. 


\section{Preliminary}
\subsection{Diffusion and Flow Models}
Diffusion models~\cite{diffusion,diffusion2,ddpm} learn to reverse a predefined forward noising process, transforming a simple prior distribution into a complex data distribution $p_\text{data}$.
The forward process gradually adds noise to a clean sample $\vx_0 \sim p_\text{data}$ according to:
\begin{equation}
    \vx_t = \alpha_t \vx_0 + \sigma_t \epsilonv, \quad \text{where} \quad \epsilonv \sim \gN(\bm{0}, \rmI).
\end{equation}
The functions $\alpha_t$ and $\sigma_t$ define a noise schedule that maps a clean sample $\vx_0$ at $t=0$ to a noisy sample $\vx_1$ approximating the prior at $t=1$.
To learn the reverse process, Flow Matching~\cite{flow1, flow2} trains a time-dependent vector field $\vv_\theta(\vx_t, t)$ by minimizing a loss function that matches it to the target velocity field of the forward process:
\begin{equation}
\label{eq:flow_matching_loss}
    \E_{t, \vx_0, \epsilonv} \left[ w(t) \left\| \vv_\theta(\alpha_t \vx_0 + \sigma_t \epsilonv, t) - (\dot{\alpha}_t \vx_0 + \dot{\sigma}_t \epsilonv) \right\|_2^2 \right],
\end{equation}
where $w(t)$ is a weighting function and $\dot{f}_t$ denotes the time derivative $\mathrm{d}f_t/\mathrm{d}t$.
Once trained, the learned field $\vv_\theta$ governs the generative process via a probability flow Ordinary Differential Equation (ODE)~\cite{sde}:
\begin{equation}
\label{eq:generative_ode}
    \frac{\mathrm{d}\vx_t}{\mathrm{d}t} = \vv_\theta(\vx_t, t).
\end{equation}
New samples are generated by numerically integrating this ODE backward in time from $t=1$ to $t=0$, starting from a random sample $\vx_1$ drawn from the prior.

\subsection{One-Step Denoising for Reward Evaluation}

A core challenge in aligning diffusion models~\cite{ddpm} is that optimization occurs on noisy latents $\vx_t$, whereas reward models require a clean image $\vx_0$ for evaluation. A common approach \cite{refl,draft} is to perform one-step denoising to predict an image $\hat{\vx}_0$ from $\vx_t$, but this prediction is often inaccurate for early, high-noise timesteps (large $t$).

To overcome this instability, we adopt the ground-truth noise prior technique~\cite{srpo,diffusionnft}. We begin with a clean image $\vx_0$ and a known noise sample $\epsilonv_{\text{gt}}$ to create a specific noisy state $\vx_t = \alpha_t \vx_0 + \sigma_t \epsilonv_{\text{gt}}$. The model then predicts the noise $\epsilonv_\theta(\vx_t, t)$, which allows us to reconstruct a high-fidelity, differentiable estimate of the clean image for reward evaluation:
\begin{equation}
\label{eq:one_step_denoise}
    \hat{\vx}_0 = \frac{\vx_t - \sigma_t \epsilonv_\theta(\vx_t, t)}{\alpha_t}.
\end{equation}
This reliable reconstruction of $\hat{\vx}_0$ provides a stable reward signal across all timesteps, enabling us to sample $t$ uniformly from $[0, 1]$ for robust training.

\section{Methodology}
Aligning T2I models with human preference aims to produce aesthetically appealing and diverse visual content.
RL algorithms accomplish this by optimizing the model to maximize a predefined reward score. However, this process is prone to reward hacking, making its mitigation a critical challenge. 
In this section, we first analyze the challenge of reward hacking, identifying that the neglect of diversity during alignment leads to \textit{Preference Mode Collapse}. Subsequently, to address this phenomenon, we propose \textit{\ours}, our novel framework. Finally, we introduce \textit{\bench}, a benchmark specifically for evaluating diversity.


\subsection{Preference Mode Collapse}

While reinforcement learning successfully aligns T2I models with human preference, it is prone to reward hacking. Existing methods mitigate this by preventing quality degradation, but overlook a critical side effect: \textit{a sharp decline in generative diversity}. As illustrated in Fig.~\ref{fig: teaser}, this creates a problematic trade-off where models produce homogeneous outputs despite high reward scores. We identify this phenomenon—reward hacking from a diversity perspective—as Preference Mode Collapse (PMC).

The reasons for this oversight are twofold. First, evaluating diversity is considerably more challenging than quality. Unlike quality, evaluating diversity in T2I alignment remains an under-explored challenge. Defining standardized metrics is difficult, and existing ones are often too computationally expensive to be used as direct reward signals, thus hindering explicit regularization.
Second, and more fundamentally, any reward model possesses intrinsic preferences. The optimization process naturally drives the T2I model to overfit these preferences, inevitably causing the generative distribution to "collapse" towards a monolithic style favored by the reward model and lose diversity.
Therefore, \textit{mitigating these inherent model biases to obtain a more faithful reward signal} is a viable strategy to address this collapse.

To effectively mitigate PMC, it must first be quantitatively measured. We thus introduce \bench (Sec.~\ref{sec:bench}), a new benchmark to quantify generative diversity. 
Building on this, we propose \ourshort, a novel framework designed to mitigate the reward model's intrinsic biases and prevent this collapse.

\begin{figure}[!t]
  \centering
    \includegraphics[width=1.0\linewidth]{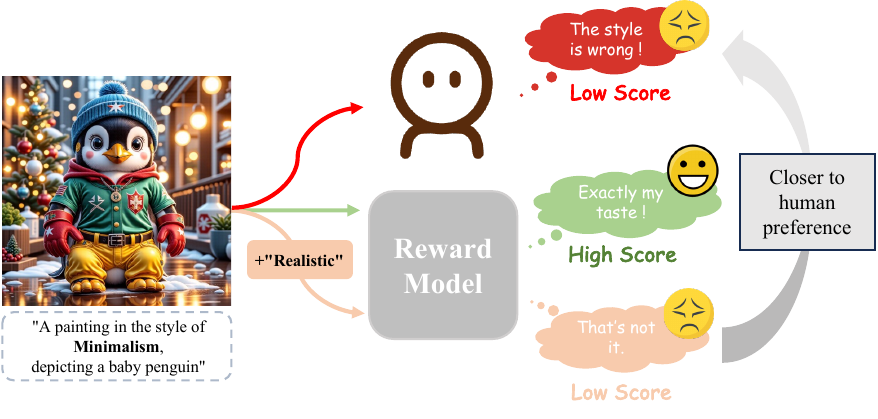}
    \caption{
      \textbf{Correcting the Reward Signal via Prompt Perturbation}. 
      An image generated for a \textbf{minimalism} prompt is instead \textbf{oily} and \textbf{overly-rendered}. 
      This style mismatch is identified by a human (low score), but the reward model assigns a high score due to its intrinsic bias. We counteract this by perturbing the prompt with descriptors like ''Realistic'' to produce a more accurate reward signal aligned with human preference.
    }
    \vspace{-10pt}
    \label{fig: demo}
\end{figure}

\begin{figure*}[!t]
  \centering
    \includegraphics[width=1.0\textwidth]{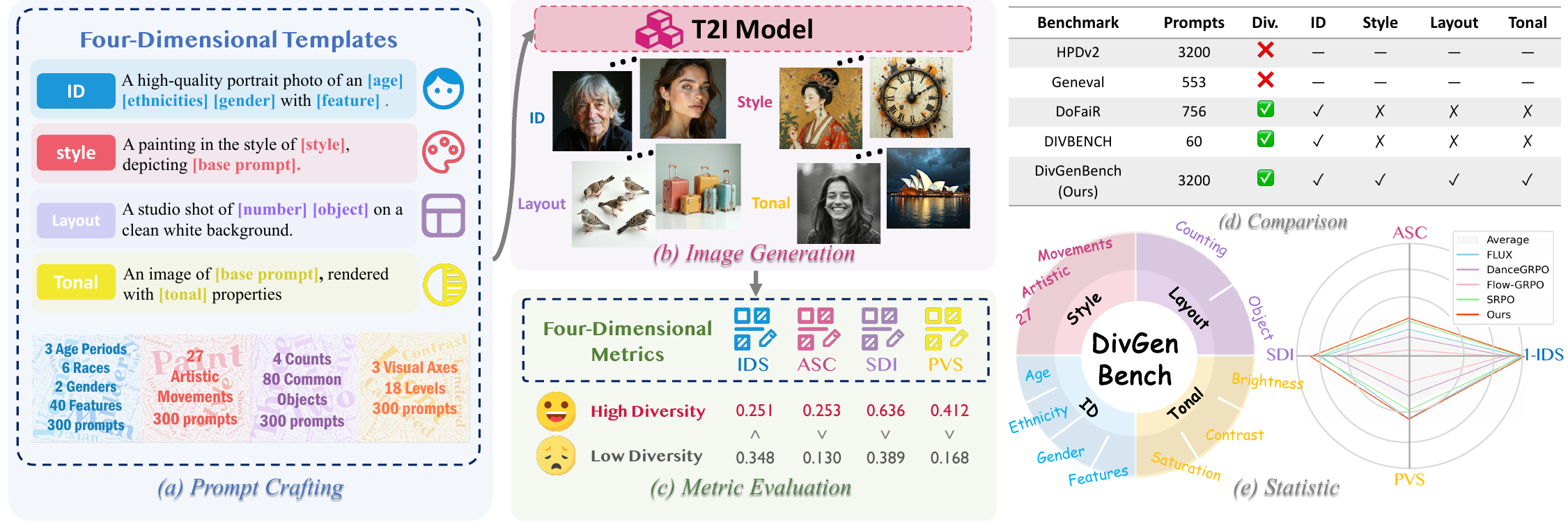}
    \caption{\textbf{Overview Construction and Evaluation Pipeline of Our \bench.} (a) A systematic process for prompt construction. (b) Generation of images across four distinct  dimensions based on different prompts. (c) Quantitative evaluation using our four proposed metrics: Identity Divergence Score (IDS), Artistic Style Coverage (ASC), Spatial Dispersion Index (SDI), and Photographic Variance Score (PVS). (d) A comparative analysis positioning our benchmark against existing ones. (e) Detailed benchmark statistics and performance comparison of state-of-the-art methods.}
    \label{fig: bench}
\end{figure*}

\subsection{Directional Decoupling Alignment}
In the standard alignment process, a sampled image and its corresponding prompt are fed into a reward model, which outputs a feedback signal for gradient updates. 
However, as noted in \cite{dancegrpo,srpo,flowgrpo}, reward models such as HPS-v2.1~\cite{hps} can cause the T2I model to generate overly smooth and glossy images. This phenomenon leads to inflated reward scores, which genuine human evaluators would not rate as highly.
As illustrated in Fig.~\ref{fig: demo}, we observe that this bias can be counteracted by directionally perturbing the prompt. For instance, when we append the descriptor "realistic" to the prompt of a glossy image, the corresponding reward is suppressed, better aligning the feedback with human aesthetic judgment. This demonstrates the efficacy of a corrected reward signal.
While perturbing prompts with manually selected words can yield a bias-suppressed reward, this approach is inherently limited, as the vocabulary space is discrete and pre-defining such words is inefficient. This limitation motivates our strategy to \textit{identify a direction in the continuous prompt embedding space that can systematically adjust the reward signal}. 
By operating within the continuous embedding space of the text encoder, we can apply a more principled correction to mitigate the biases of the reward model.

As illustrated in Fig.~\ref{fig: pipeline}, \ourshort is a two-stage framework that addresses the challenge by decoupling the process into two distinct steps: \textit{reward signal correction} and \textit{guided alignment}. 
In Stage 1, we empirically identify a direction within the continuous embedding space to create a corrected reward signal. 
In Stage2, we leverage this learned direction to guide the generator's gradient updates, preventing it from over-optimizing into specific modes.

\begin{figure}[!t]
  \centering
    \includegraphics[width=1.0\linewidth]{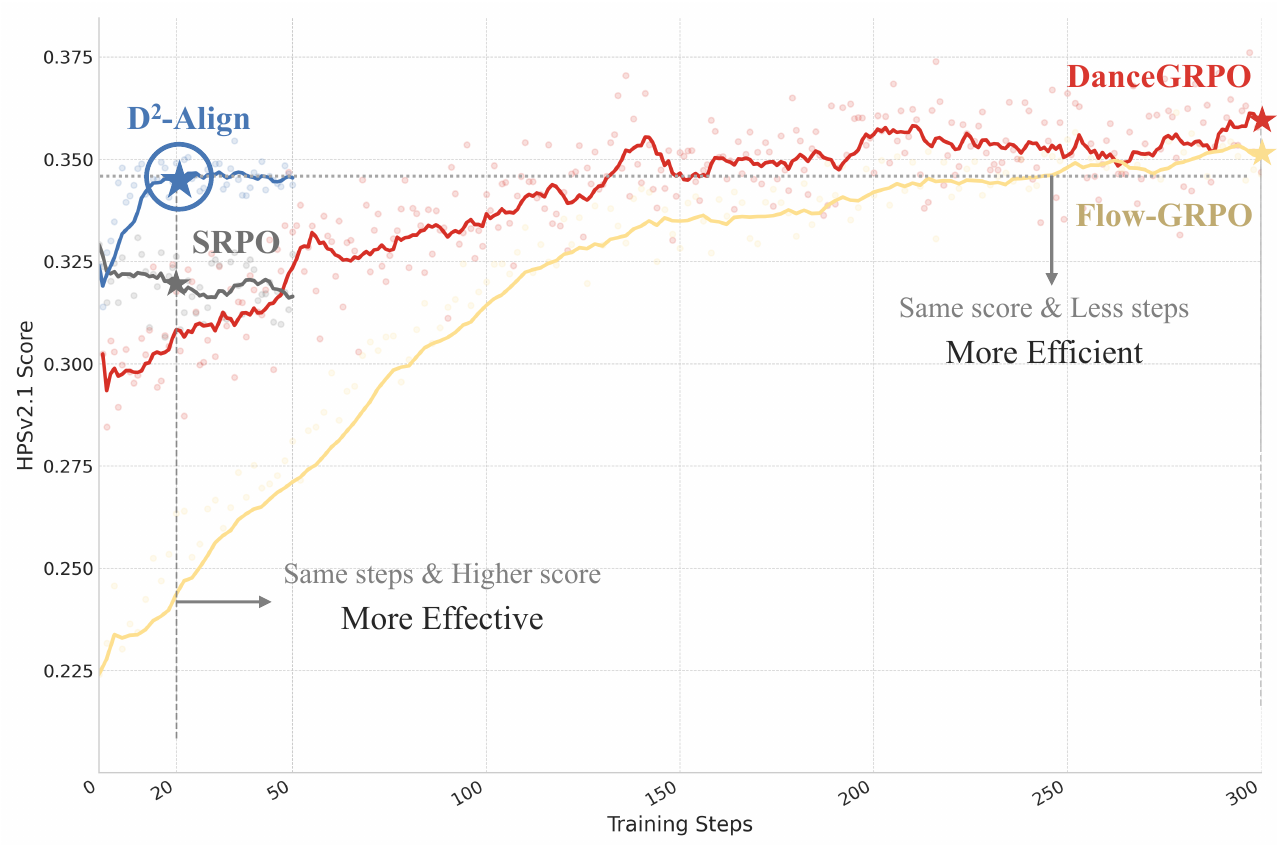}
    \caption{
        \textbf{Training Efficiency and Effectiveness Comparison}. 
        \ourshort outperforms baselines by being both more effective and more efficient. 
        It achieves a higher score in fewer steps, whereas methods like DanceGRPO and Flow-GRPO require over 250 steps to attain a similar level of performance.
    }
    \vspace{-20pt}
    \label{fig: train}
\end{figure}

\textbf{Stage 1.} Following \cite{refl, dancegrpo,diffusionnft,mixgrpo,prefgrpo}, we consider HPS-v2.1~\cite{hps} as our reward model, which is a CLIP-based model finetuned on human preference data.
The original reward score $R(\vx_0, c)$ is computed by a scoring function:
\begin{equation}
\label{eq:original_reward}
    R(\vx_0, c) = \score(\Phi_\text{img}(\vx_0), \Phi_\text{text}(c))
\end{equation}
where the $\score$ function computes the cosine similarity between the input embeddings, and $\Phi_\text{img}$ and $\Phi_\text{text}$ are the image and text encoders of the model, respectively.
We freeze the generator $G_\theta$ and introduce a learnable vector $\boldsymbol{b}_v \in \mathbb{R}^d$. 
Inspired by \cite{cfg,reneg}, we use this vector to construct a guided reward signal. First, we define two perturbed embeddings, $\boldsymbol{e}_{+}$ and $\boldsymbol{e}_{-}$, from the original text embedding $\boldsymbol{e}_{\text{text}} = \Phi_\text{text}(c)$:
\begin{align}
    \boldsymbol{e}_{+} &= \normalize(\boldsymbol{e}_{\text{text}} + \boldsymbol{b}_v) \\
    \boldsymbol{e}_{-} &= \normalize(\boldsymbol{e}_{\text{text}} - \boldsymbol{b}_v)
\end{align}
We then construct a new, guided text embedding $\tilde{\boldsymbol{e}}_{\text{text}}$ that extrapolates from the negative direction towards the positive one, controlled by a guidance scale $\omega > 1$:
\begin{equation}
\label{eq:guided_embedding}
    \tilde{\boldsymbol{e}}_{\text{text}} = \boldsymbol{e}_{-} + \omega \cdot (\boldsymbol{e}_{+} - \boldsymbol{e}_{-})
\end{equation}
The guided reward, $R_{\text{guided}}$, is computed by applying the same scoring function to the guided text embedding:
\begin{equation}
    R_{\text{guided}}(\vx_0, c; \boldsymbol{b}_v) = \score(\boldsymbol{e}_\text{img}, \tilde{\boldsymbol{e}}_{\text{text}})
\end{equation}
To provide a clean image $\vx_0$ for the reward calculation, we leverage the ground-truth noising strategy. Specifically, we add a known noise $\epsilonv \sim \mathcal{N}(0, \mathbf{I})$ to a ground-truth image according to the forward process $\vx_t = \alpha_t \vx_0 + \sigma_t \epsilonv$ to obtain a noisy latent $\vx_t$. The original image $\vx_0$ is then recovered using the one-step denoising process described in Eq.~\eqref{eq:one_step_denoise}. 
As illustrated in Fig.~\ref{fig: train}, this approach ensures an efficient and effective training workflow.
The vector $\boldsymbol{b}_v$ is optimized by minimizing the loss $\mathcal{L}_{\text{stage1}}(\boldsymbol{b}_v) = \E_{c, \vx_0 \sim G_{\theta_{\text{frozen}}}} [-R_{\text{guided}}]$. 
We empirically verify that this learned directional vector provides a superior mechanism for reward correction compared to using hand-picked, discrete vocabulary (detailed in Sec.~\ref{sec:ablation}).


\begin{figure*}[!t]
  \centering
    \includegraphics[width=1.0\textwidth]{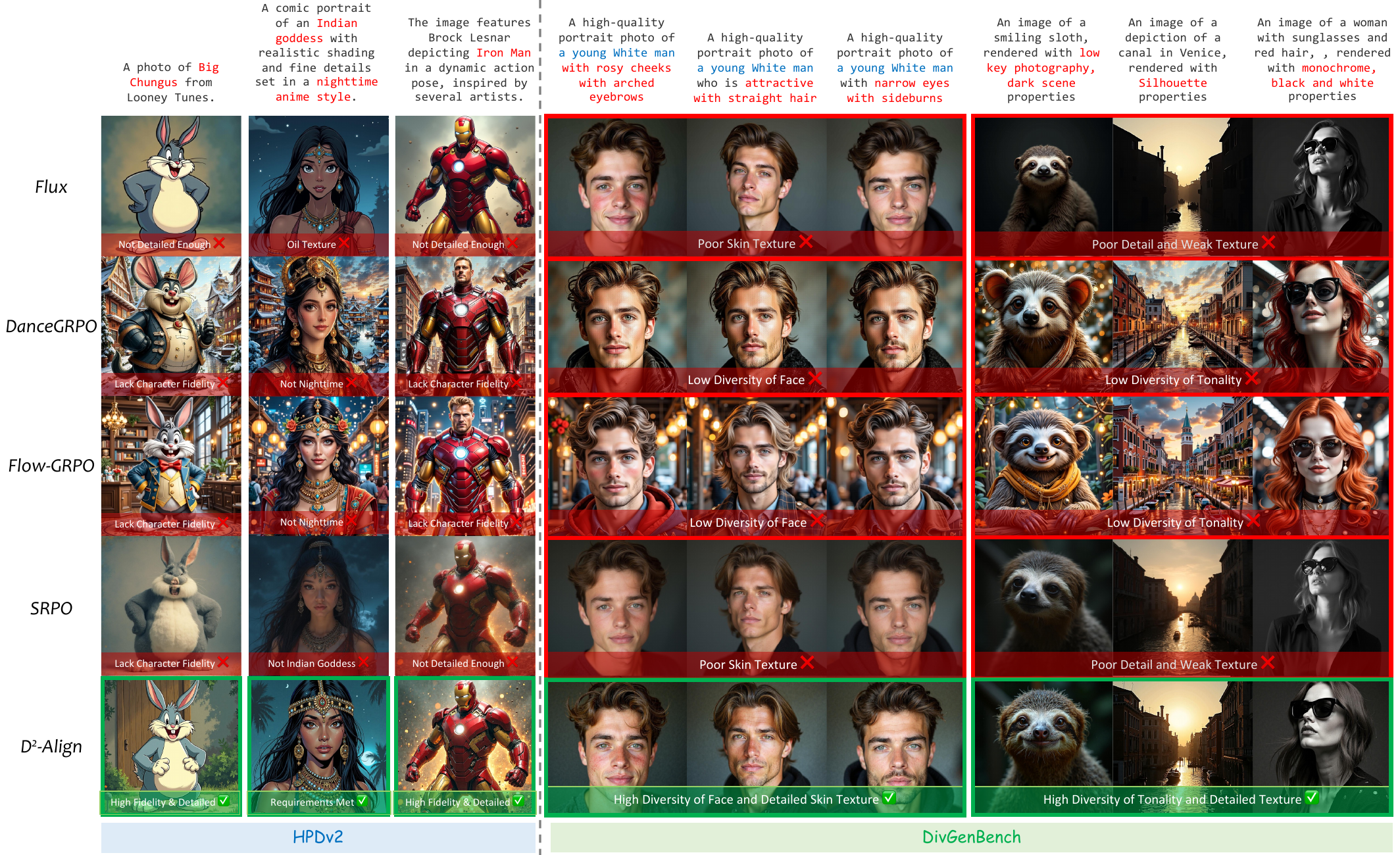}
    
    \caption{\textbf{Qualitative Comparison of \ourshort against Baselines on the HPDv2 and Our \bench Benchmarks.} \ourshort demonstrates
superior performance fidelity and diversity, overcoming the concept forgetting and stylistic limitations seen in other methods.}
    \vspace{-10pt}
    \label{fig: qualitative}
\end{figure*}

\textbf{Stage 2.} In this stage, we proceed with the alignment of the generator $G_\theta$. Let $\boldsymbol{b}_v^*$ denote the directional vector learned and frozen from Stage 1. 
The generator is then optimized by minimizing the following loss, which applies the guided reward function with the frozen vector $\boldsymbol{b}_v^*$:
\begin{equation}
    \mathcal{L}_{\text{stage2}}(\theta) = \E_{c \sim \mathcal{D}, \vx_0 \sim G_\theta(c)} [-R_{\text{guided}}(\vx_0, c; \boldsymbol{b}_v^*)]
    \label{eq:stage2_loss} 
\end{equation}
By incorporating the frozen vector $\boldsymbol{b}_v^*$, the optimization objective is fundamentally altered. Instead of naively maximizing the original reward, which is susceptible to the reward model's intrinsic biases, the generator is now guided by a corrected reward signal.
This corrected signal provides a more faithful representation of true human preference by suppressing the reward inflation often caused by these biases. Optimizing for this more credible reward naturally encourages the generator to produce outputs of higher fidelity.
Moreover, by preventing the generator from converging on specific patterns that merely cater to the reward model's biases, our approach forces it to explore a wider range of solutions, thereby preserving generative diversity and enhancing output quality.

\begin{table*}[!t]
\centering
\caption{\textbf{Comprehensive Quantitative Evaluation}. We compare FLUX and advanced methods from
the combined perspectives of human preference alignmen, semantic consistency and accuracy, showcasing performance under two distinct reward configurations: HPS-v2.1 and HPS-v2.1 + CLIP. \textbf{Ranking is performed independently for each reward configuration among the RL-based methods.} The highest score is shown in \textbf{bold}, and the second-highest score is \underline{underlined}.}
\label{tab:quantitative_evaluation_combined}
\resizebox{\textwidth}{!}{ 
\begin{tabular}{l *{8}{c}} 
\toprule
\multicolumn{1}{l|}{\multirow{3}{*}{\textbf{Method}}} & \multicolumn{5}{c|}{\textbf{Human Preference Alignment}} & \multicolumn{3}{c}{\textbf{Semantic Consistency \& Accuracy}} \\
\cmidrule(lr){2-6} \cmidrule(lr){7-9}
\multicolumn{1}{l|}{} & {\makecell{Aesthetic $\uparrow$}} & {\makecell{ImageReward $\uparrow$}} & {\makecell{Pick Score $\uparrow$}} & {\makecell{Q-Align $\uparrow$}} & \multicolumn{1}{c|}{\makecell{HPS-v2.1 $\uparrow$}} & {\makecell{CLIP $\uparrow$}} & {\makecell{DeQA $\uparrow$}} & \multicolumn{1}{c}{\makecell{GenEval $\uparrow$}} \\
\midrule
\multicolumn{1}{l|}{FLUX}             & 6.417 & 1.670 & 0.240 & 4.922 & 0.310 & 0.315 & 4.456 & 0.663 \\
\midrule
\multicolumn{9}{c}{\textit{Reward Model: HPS-v2.1}} \\
\midrule
\multicolumn{1}{l|}{DanceGRPO \cite{dancegrpo}}  & 6.068 & 1.664 & \underline{0.241} & \underline{4.930} & \underline{0.361} & 0.293 & 4.400 & 0.522 \\
\multicolumn{1}{l|}{Flow-GRPO \cite{flowgrpo}}  & 5.888 & \underline{1.703} & 0.239 & \textbf{4.969} & \textbf{0.367} & 0.283 & \underline{4.432} & 0.517 \\
\multicolumn{1}{l|}{SRPO \cite{srpo}}           & \textbf{6.614} & 1.533 & \underline{0.241} & 4.866 & 0.296 & \underline{0.302} & 4.357 & \underline{0.623} \\
\multicolumn{1}{l|}{\textbf{Ours}}       & \underline{6.450} & \textbf{1.771} & \textbf{0.246} & \textbf{4.969} & 0.343 & \textbf{0.323} & \textbf{4.484} & \textbf{0.636} \\
\midrule
\multicolumn{9}{c}{\textit{Reward Model: HPS-v2.1 + CLIP}} \\
\midrule
\multicolumn{1}{l|}{DanceGRPO \cite{dancegrpo}}        & 6.030 & 1.520 & 0.236 & \underline{4.962} & \underline{0.333} & 0.286 & 4.423 & 0.581 \\
\multicolumn{1}{l|}{Flow-GRPO \cite{flowgrpo}}        & 6.060 & \underline{1.744} & 0.239 & 4.950 & \textbf{0.343} & \underline{0.317} & 4.454 & 0.636 \\
\multicolumn{1}{l|}{SRPO \cite{srpo}}             & \underline{6.394} & 1.616 & \underline{0.240} & 4.958 & 0.310 & 0.309 & \underline{4.495} & \underline{0.659} \\
\multicolumn{1}{l|}{\textbf{Ours}}       & \textbf{6.671} & \textbf{1.762} & \textbf{0.246} & \textbf{4.970} & 0.314 & \textbf{0.328} & \textbf{4.498} & \textbf{0.660} \\
\bottomrule
\end{tabular}
} 
\vspace{-10pt}
\end{table*}

\subsection{\bench}
\label{sec:bench}

To quantify Preference Mode Collapse (PMC), we introduce \bench, a benchmark designed to evaluate a model's \textit{Generative Breadth}—its ability to follow diverse, explicit instructions. Existing benchmarks are inadequate for diagnosing PMC, as they typically largely prioritize fidelity~\cite{hps,ghosh2023geneval,t2icombench,drawbench}, or measure output variance from ambiguous prompts\cite{teotia2025dimcim}, or lack comprehensive dimensions\cite{friedrich2025beyond} and metrics\cite{orw,flowgrpo} for diversity.

\bench addresses these gaps with a "keyword-driven" prompt design that actively probes a model's generative boundaries. Its hierarchical structure encompasses four key dimensions: ID (high-level semantics), Style (mid-level aesthetics), Layout (structure and relations), and Tonal (low-level physics). 
We systematically constructed our \bench of \textbf{3,200} prompts, each dimension comprising 800 prompts, which are constructed by augmenting a set of base templates—variously derived from ~\cite{fairface,parti,coco}—with explicit attribute keywords sourced from ~\cite{CelebA,wikiart} and other custom-designed keywords.

Recognizing that a single metric is insufficient, \bench employs a suite of dimension-customized metrics for robust evaluation. We combine low-level image features with domain-specific extractors~\cite{arcface,csd,groundingdino}. We then apply our four bespoke metric calculations to compute the final scores: the Identity Divergence Score (IDS), Artistic Style Coverage (ASC), Spatial Dispersion Index (SDI), and Photographic Variance Score (PVS), inspired by~\cite{irs,apg}. Detailed metric designs are provided in \supp

\section{Experiment}
\subsection{Implementation Details}
We evaluate our method against several leading RL alignment baselines on the state-of-the-art T2I model, FLUX.1.Dev~\cite{flux2024} . The primary reward signal is the HPS-v2.1 trained on the Human Preference Dataset (HPD) v2~\cite{hps}.

Following DanceGRPO~\cite{dancegrpo}, we also conduct experiments with a multi-reward combination of HPS-v2.1 and CLIP Score~\cite{hessel2021clipscore} for a more comprehensive comparison
Our evaluation is comprehensive, assessing model performance from two key perspectives: \textit{quality and diversity}.
For quality assessment, we employ a suite of metrics. Aesthetic appeal is measured using state-of-the-art predictors~\cite{qalign, aes, refl, pickscore}.
To evaluate text-image alignment, we use CLIP Score. Furthermore, we assess semantic fidelity using~\cite{ghosh2023geneval}. For diversity evaluation, we utilize \bench. 

\subsection{Qualitative Evaluation}
The qualitative comparisons on HPDv2 and our \bench are presented in Fig.~\ref{fig: qualitative}. Our method consistently outperforms the baselines in terms of fidelity, text-to-image alignment and diversity.

\begin{figure*}[!t]
  \centering
    \includegraphics[width=1.0\textwidth]{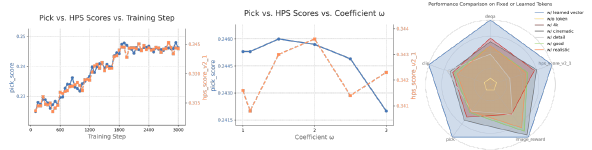}
    
    \caption{\textbf{Ablation Studies on the Key Components and Hyperparameters of \ourshort.}
\textbf{(Left)} Analysis of the $\mathbf{b}_{v}$ directional vector's convergence. 
\textbf{(Middle)} Sensitivity analysis of the guidance scale $\omega$. 
\textbf{(Right)} Effectiveness of the learned directional vector. The radar chart compares our full method (using the learned $\mathbf{b}_{v}$) against variants using manually-defined discrete tokens (e.g., "realistic") and no directional correction.}
    \vspace{-15pt}
    \label{fig: ablation}
\end{figure*}

On HPDv2 (Cols. 1-3), baselines prone to PMC, such as DanceGRPO, FlowGRPO and SRPO, exhibit significant concept forgetting, failing to render well-known subjects like "Big Chungus" (Col. 1) and "Iron Man" (Col. 3). They also fail on complex semantic consistency, unable to synthesize all attributes for the "Indian goddess... nighttime anime style" prompt (Col. 2), which our method achieves.
This degradation is magnified on our \bench (Cols. 4-8). DanceGRPO and Flow-GRPO suffer severe PMC, generating near-identical faces despite prompts requiring ID diversity (Cols. 4-6). Similarly, they fail on tonal diversity (Cols. 7-8), defaulting to a homogeneous aesthetic instead of the requested "low key" or "black and white" styles.
In contrast, our method not only preserves these concepts but also demonstrates superior fidelity and detail over other baselines like FLUX and SRPO. Our method successfully balances high fidelity with robust instruction-following capabilities. More qualitative results can be found in \supp

\subsection{Quantitative Evaluation}
We quantitatively evaluate the model's performance from the combined perspectives of quality and diversity.
From a quality perspective, as shown in Tab.~\ref{tab:quantitative_evaluation_combined}, our method achieves superior performance on key metrics. This indicates that our approach successfully aligns with human preference while maintaining high fidelity. Furthermore, our method also outperforms all baselines in text-to-image alignment and semantic consistency, demonstrating its robust instruction-following capabilities.
In terms of diversity, Tab.~\ref{tab:quantitative_evaluation_diversity_single_col} shows that our method attains the highest scores on the \bench. This result provides strong evidence that our proposed method effectively prevents the model from over-optimizing towards the reward model's preferred modes, thereby mitigating the PMC. 
It is particularly noteworthy that DanceGRPO and FlowGRPO, which achieve artificially high scores on HPS-v2.1, exhibit a significant drop in diversity. 
This finding further validates the necessity of our approach and highlights the limitations of relying solely on a single reward score for evaluation. 
To provide a measure of alignment with human preference, we conducted a user study following~\cite{dancegrpo}. \ourshort achieved the best win rate (see \supp for details).

\begin{table}[htb]
\centering
\caption{\textbf{Quantitative Evaluation of Generative Diversity on \bench}. We compare the FLUX and other advanced methods, showcasing performance under two distinct reward configurations and four metrics, i.e., Identity Divergence Score (IDS), Artistic Style Coverage (ASC), Spatial Dispersion Index (SDI), and Photographic Variance Score (PVS). \textbf{Ranking is performed independently for each reward configuration among the RL-based methods}. The best score is shown in \textbf{bold}, and the second-best score is \underline{underlined}.
}
\label{tab:quantitative_evaluation_diversity_single_col}
\begin{tabular}{l *{4}{c}}
\toprule
\multicolumn{1}{l|}{\textbf{Method}} & {\makecell{IDS $\downarrow$}} & {\makecell{ASC $\uparrow$}} & {\makecell{SDI $\uparrow$}} & {\makecell{PVS $\uparrow$}} \\
\midrule
\multicolumn{1}{l|}{FLUX} & 0.280 & 0.179 & 0.563 & 0.408 \\
\midrule
\multicolumn{5}{c}{\textit{Reward Model: HPS-v2.1}} \\
\midrule
\multicolumn{1}{l|}{DanceGRPO} & 0.348 & 0.130 & 0.488 & 0.259 \\
\multicolumn{1}{l|}{Flow-GRPO} & 0.391 & 0.044 & 0.389 & 0.168 \\
\multicolumn{1}{l|}{SRPO} & \underline{0.259} & \underline{0.234} & \underline{0.580} & \underline{0.352} \\
\multicolumn{1}{l|}{\textbf{Ours}} & \textbf{0.251} & \textbf{0.253} & \textbf{0.636} & \textbf{0.412} \\
\midrule
\multicolumn{5}{c}{\textit{Reward Model: HPS-v2.1 + CLIP}} \\
\midrule
\multicolumn{1}{l|}{DanceGRPO} & \underline{0.252} & 0.108 & 0.523 & 0.344 \\
\multicolumn{1}{l|}{Flow-GRPO} & 0.276 & 0.133 & 0.506 & 0.316 \\
\multicolumn{1}{l|}{SRPO} & 0.306 & \underline{0.198} & \underline{0.559} & \underline{0.402} \\
\multicolumn{1}{l|}{\textbf{Ours}} & \textbf{0.237} & \textbf{0.247} & \textbf{0.631} & \textbf{0.418} \\
\bottomrule
\end{tabular}
\vspace{-10pt}
\end{table}

\subsection{Ablation Study}
\label{sec:ablation}
In this section, we conduct a comprehensive ablation study on \ourshort to validate its effectiveness and robustness.

\textbf{Convergence Analysis of the Learned Vector ($\mathbf{b}_{v}$).}
To analyze the convergence of the directional vector, we evaluated the performance of $\mathbf{b}_{v}$ sampled from different training steps.
As depicted in Fig.~\ref{fig: ablation} (left), the corrective effect of $\mathbf{b}_{v}$ becomes evident and robust after approximately 2000 training steps, yielding significant performance improvements from that point onward.

\textbf{Ablation on the Guidance Scale ($\omega$).}
To investigate the impact of the guidance scale $\omega$, we evaluate the optimization performance using the learned directional vector $\mathbf{b}_{v}$ under various scale settings. 
We empirically set $\omega$ to 1.5, as this value yields superior results on both HPS-v2.1 and Pickscore, as demonstrated in Fig.~\ref{fig: ablation} (middle).

\textbf{Effectiveness of the Learned Vector ($\mathbf{b}_{v}$).}
To validate that our learned continuous vector, $\mathbf{b}_{v}$, more effectively mitigates over-optimization towards high rewards. 
We contrast our method against a baseline that uses manually selected discrete words representing fidelity and aesthetics.
As illustrated in Fig.~\ref{fig: ablation} (right), our approach consistently outperforms this baseline of hand-picked words across all evaluated metrics. 
Furthermore, compared to using the uncorrected reward signal, our method demonstrates a comprehensive improvement across the board.

\textbf{Generalizability of the Learned Vector ($\mathbf{b}_{v}$).}
To evaluate the generalizability of our learned $\mathbf{b}_{v}$, we applied it as a corrective signal to other existing methods that are susceptible to PMC. 
The results show that even when integrated into these external frameworks, $\mathbf{b}_{v}$ consistently mitigates the collapse by achieving a more favorable balance between fidelity and diversity. 
A detailed analysis of these cross-method experiments is provided in \supp


\section{Conclusion}

We identified Preference Mode Collapse (PMC), 
a reward hacking from a diversity perspective.
We proposed \ourshort, a novel framework that counteracts PMC by mitigating the reward model's intrinsic biases, thus breaking the common trade-off between fidelity and diversity. To measure this phenomenon, we introduced \bench, a new diversity-centric benchmark. Our results show that \ourshort achieves state-of-the-art performance, simultaneously improving both human preference scores and generative diversity.

{
    \small
    \bibliographystyle{ieeenat_fullname}
    \bibliography{main}
}

\clearpage
\maketitlesupplementary
\label{sec:appendix}

\renewcommand{\thesection}{\Alph{section}}
\setcounter{section}{0} 
\renewcommand{\thesection}{\Alph{section}}
\renewcommand{\thesubsection}{\Alph{section}.\arabic{subsection}}
\renewcommand{\thesubsubsection}{\Alph{section}.\arabic{subsection}.\arabic{subsubsection}}

\section{\ourshort Algorithm}
\label{sec:algorithm}
The \ourshort framework is a two-stage process that decouples reward signal correction from policy alignment. The detailed process for each stage is provided in \cref{alg:d2_align_s1} and \cref{alg:d2_align_s2}.

\textbf{Stage 1 (Algorithm \ref{alg:d2_align_s1})} focuses on learning a directional correction vector $\boldsymbol{b}_v$. In this stage, the policy model $G_{\theta}$ is frozen, and only the vector $\boldsymbol{b}_v$ is optimized to create a guided reward signal that corrects for reward model biases.

\textbf{Stage 2 (Algorithm \ref{alg:d2_align_s2})} then performs the guided alignment of the policy model itself. The optimized vector $\boldsymbol{b}_v^*$ from Stage 1 is frozen, and the policy model $G_{\theta}$ is unfrozen. The generator's parameters $\theta$ are then updated by optimizing for the guided reward signal defined by $\boldsymbol{b}_v^*$.

\begin{algorithm}[h!]
\caption{\ourshort Stage 1: Learning Directional Correction}
\label{alg:d2_align_s1}
\begin{algorithmic}[1]
\Require Initial policy model $G_{\theta}$ (as $\epsilonv_\theta$); reward model $R$ ($\Phi_\text{img}, \Phi_{\text{text}}, \score$); prompt dataset $\mathcal{C}$; guidance scale $\omega$; Stage 1 total timesteps $T_1$ for training; diffusion coefficients $\alpha_t, \sigma$.
\Ensure Optimized directional vector $\boldsymbol{b}_v^*$.

\State Initialize learnable directional vector $\boldsymbol{b}_v \in \mathbb{R}^d$
\State Freeze generator $G_{\theta}$ ($\epsilonv_\theta$)
\For{timestep $T_1$ to $t = 1$}
\State Sample prompt $c \sim \mathcal{C}$
\State Generate clean image $\vx_0 \sim G_{\theta}(c)$
\State Get text embedding $\boldsymbol{e}_{\text{text}} \gets \Phi_{\text{text}}(c)$ 
\State Sample noise $\epsilonv_{\text{gt}} \sim \mathcal{N}(0, \mathbf{I})$
\State Create noisy latent $\vx_t \gets \alpha_t \vx_0 + \sigma_t \epsilonv_{\text{gt}}$
\State Predict noise $\epsilonv_{\text{pred}} \gets \epsilonv_\theta(\vx_t, t, c)$
\State Perform one-step ODE sampling to get $\vx_{t-1}$
\State Reconstruct $\hat{\vx}_0 \gets (\vx_{t-1} - \sigma_{t-1} \epsilonv_{\text{gt}}) / \alpha_{t-1}$
\State Get image embedding $\boldsymbol{e}_{\text{img}} \gets \Phi_{\text{img}}(\hat{\vx}_0)$

\State Calculate $\boldsymbol{e}_{+} \gets \normalize(\boldsymbol{e}_{\text{text}} + \boldsymbol{b}_v)$
\State Calculate $\boldsymbol{e}_{-} \gets \normalize(\boldsymbol{e}_{\text{text}} - \boldsymbol{b}_v)$
\State Construct guided embedding $\tilde{\boldsymbol{e}}_{\text{text}} \gets \boldsymbol{e}_{-} + \omega \cdot (\boldsymbol{e}_{+} - \boldsymbol{e}_{-})$
\State Compute guided reward $R_{\text{guided}} \gets \score(\boldsymbol{e}_\text{img}, \tilde{\boldsymbol{e}}_{\text{text}})$
\State Update $\boldsymbol{b}_v$ by minimizing $\mathcal{L}_{\text{stage1}}(\boldsymbol{b}_v) = -R_{\text{guided}}$
\EndFor
\State \Return $\boldsymbol{b}_v^* \gets \boldsymbol{b}_v$
\end{algorithmic}
\end{algorithm}

\begin{algorithm}[h!]
\caption{\ourshort Stage 2: Guided Alignment}
\label{alg:d2_align_s2}
\begin{algorithmic}[1]
\Require Policy model $G_{\theta}$ (as $\epsilonv_\theta$); reward model $R$ ($\Phi_\text{img}, \Phi_{\text{text}}, \score$); prompt dataset $\mathcal{C}$; guidance scale $\omega$; Stage 2 total timesteps $T_2$ for training; diffusion coefficients $\alpha_t, \sigma$.
\Require Frozen directional vector $\boldsymbol{b}_v^*$ (from Stage 1).
\Ensure Optimized policy model $G_{\theta^*}$.

\State Unfreeze generator $G_{\theta}$ ($\epsilonv_\theta$)
\For{timestep $T_2$ to $t = 1$}
\State Sample prompt $c \sim \mathcal{C}$
\State Generate clean image $\vx_0 \sim G_{\theta}(c)$
\State Get text embedding $\boldsymbol{e}_{\text{text}} \gets \Phi_{\text{text}}(c)$
\State Sample noise $\epsilonv_{\text{gt}} \sim \mathcal{N}(0, \mathbf{I})$
\State Create noisy latent $\vx_t \gets \alpha_t \vx_0 + \sigma_t \epsilonv_{\text{gt}}$
\State Predict noise $\epsilonv_{\text{pred}} \gets \epsilonv_\theta(\vx_t, t, c)$
\State Perform one-step ODE sampling to get $\vx_{t-1}$
\State Reconstruct $\hat{\vx}_0 \gets (\vx_{t-1} - \sigma_{t-1} \epsilonv_{\text{gt}}) / \alpha_{t-1}$
\State Get image embedding $\boldsymbol{e}_{\text{img}} \gets \Phi_{\text{img}}(\hat{\vx}_0)$

\State Calculate $\boldsymbol{e}_{+} \gets \normalize(\boldsymbol{e}_{\text{text}} + \boldsymbol{b}_v^*)$
\State Calculate $\boldsymbol{e}_{-} \gets \normalize(\boldsymbol{e}_{\text{text}} - \boldsymbol{b}_v^*)$ 
\State Construct guided embedding $\tilde{\boldsymbol{e}}_{\text{text}} \gets \boldsymbol{e}_{-} + \omega \cdot (\boldsymbol{e}_{+} - \boldsymbol{e}_{-})$
\State Compute guided reward $R_{\text{guided}} \gets \score(\boldsymbol{e}_\text{img}, \tilde{\boldsymbol{e}}_{\text{text}})$
\State Update generator $\theta$ by minimizing $\mathcal{L}_{\text{stage2}}(\theta) = -R_{\text{guided}}$
\EndFor
\State \Return $G_{\theta^*} \gets G_{\theta}$
\end{algorithmic}
\end{algorithm}



\section{Experimental Setting Details}
\label{sec:exp_setup}

We conduct all reinforcement learning experiments using FLUX.1.Dev~\cite{flux2024} as the base T2I model, following the state-of-the-art methodology. For fairness and comprehensive comparison, all alignment baselines are retrained on the Human Preference Dataset (HPD) v2~\cite{hps} using two standardized reward combinations: HPS-v2.1~\cite{hps} and HPS-v2.1~\cite{hps} + CLIP~\cite{hessel2021clipscore} Score. All experiments are conducted on NVIDIA
H20 GPUs with 96GB memory.

\subsection{Baseline Implementation Details}
\label{sec:baseline_details}
To ensure a fair comparison, we standardize the training steps for major baselines and align all reward functions. For the baseline comparisons, we follow the original implementations provided in their official repositories. Details on the training steps for each competing method are provided below.

\begin{itemize}
    \item \textbf{DanceGRPO} \cite{dancegrpo}: We follow the default open-source hyperparameters and train the model for 300 steps.
    \item \textbf{Flow-GRPO} \cite{flowgrpo}: Hyperparameters were adopted from a stable PickScore training configuration. The training length is set to 300 steps to maintain alignment with DanceGRPO.
    \item \textbf{SRPO} \cite{srpo}: The model is trained using its default configuration for the official recommended length of 20 steps.
\end{itemize}

\subsection{\ourshort Hyperparameters}
\label{sec:our_hyperparams}

Our proposed \ourshort framework employs the two-stage training approach. Stage 1 (Directional Correction) trains the correction vector $\boldsymbol{b}_v$ for 3000 steps, after which the policy model $G_{\theta}$ is aligned in Stage 2 with training 20 steps. Key hyperparameters specific to the \ourshort are listed in Tab.~\ref{tab:d2_align_params}.

\begin{table}[t]
\centering
\caption{\textbf{Comprehensive Hyperparameters of \ourshort.}}
\label{tab:d2_align_params}
\resizebox{\linewidth}{!}{
    \setlength{\tabcolsep}{3pt} 
    \begin{tabular}{l l c}
    \toprule
    \textbf{Category} & \textbf{Parameter} & \textbf{Value} \\
    \midrule
    \multirow{6}{*}{\textbf{GENERAL \& MODEL}}
    & Random Seed & 42 \\ 
    & Resolution (H $\times$ W) & $720 \times 720$ \\ 
    & Mixed Precision & bf16 \\ 
    & Gradient Checkpointing & Enabled \\ 
    & Dataloader Workers & 4 \\ 
    & Use EMA & Disabled \\ 
    \midrule
    \multirow{6}{*}{\textbf{OPTIMIZATION}}
    & Optimizer & AdamW \\ 
    & Learning Rate & \num{5e-6} \\ 
    & Weight Decay & \num{1e-4} \\ 
    & Gradient Clip Norm & 0.1 \\ 
    & LR Warmup Steps & 0 \\ 
    & Gradient Accumulation Steps & 2 \\ 
    \midrule
    \multirow{3}{*}{\textbf{RL ALIGNMENT}}
    & $\omega$ & 1.5 \\ 
    & Stage 1 ($b_v$) Max Steps & 3000 \\ 
    & Stage 2 ($G_{\theta}$) Max Steps & 20 \\ 
    \midrule
    \multirow{5}{*}{\textbf{INFERENCE}}
    & Sampling Steps & 25 \\ 
    & Inference Steps & 50 \\ 
    & Train Guidance & 1.0 \\ 
    & Shift & 3 \\ 
    & Train Batch Size & 1 \\ 
    & SP Size / Train SP Batch Size & 1 / 1 \\ 
    \bottomrule
    \end{tabular}
}
\vspace{-10pt} 
\end{table}

\begin{table*}[t]
\centering
\caption{\textbf{Prompt Construction Templates and Examples for Each Dimension of \bench.} Brackets indicate keywords sampled from curated attribute lists.}
\label{tab:prompt_examples}
\small 
\begin{tabularx}{\textwidth}{l X X}
\toprule
\textbf{Dimension} & \textbf{One of Templates} & \textbf{Example} \\
\midrule
ID & A high-quality portrait photo of an [age] [ethnicities] [gender] with [feature]. & A high-quality portrait photo of an elderly South Asian woman with arched eyebrows wearing a necklace. \\
\addlinespace 
Style & A painting in the style of [style], depicting [base prompt]. & A painting in the style of Rococo, depicting a silver fire hydrant next to a sidewalk. \\
\addlinespace
Layout & A studio shot of [number] [object] on a clean white background. & A studio shot of three boats on a clean white background. \\
\addlinespace
Tonal & An image of [base prompt], rendered with [tonal] properties. & An image of ten children on a couch, rendered with dimly lit properties. \\
\bottomrule
\end{tabularx}
\end{table*}

\section{\bench Construction and Metrics}
\label{sec:DivGenBench_details}
\subsection{Prompt Construction and Examples}
Our \textbf{3,200} "keyword-driven" prompts are systematically generated using distinct templates for each of the four dimensions, as summarized in Tab.~\ref{tab:prompt_examples}. Each dimension uses a specific templating strategy to augment base content with explicit attribute keywords.
\begin{itemize}
    \item \textbf{ID}: Motivated by \cite{fairface}, prompts are built using the template: "A high-quality portrait photo of a/an [age] [ethnicity] [gender] [features]". We combine 3 ages, 6 ethnicities, 2 genders, and 40 physical features (e.g., "with arched eyebrows") derived from CelebA\cite{CelebA}. A conflict-resolution mechanism ensures logical coherence.
    \item \textbf{Style}: We pair 27 classic art styles from WikiArt\cite{wikiart} with base content prompts from Parti\cite{parti}. 
    \item \textbf{Layout}: We combine 80 COCO\cite{coco} object classes with 4 counts (two, three, four, five) using designed template to test numerical control and spatial diversity.
    \item \textbf{Tonal}: We augment Parti\cite{parti} base prompts with 18 fine-grained keywords across 3 sub-dimensions: Saturation, Contrast, and Brightness.
\end{itemize}

\subsection{Metric Calculation Details}
To quantify the extent of PMC, we employ four dimension-customized metrics that measure the model's generative breadth.

\noindent\textbf{Identity Divergence Score (IDS):} We use ArcFace\cite{arcface} to extract a 512-D identity embedding $v_i$ for each of $N$ generated faces. The score is the average pairwise cosine similarity between all unique identity vectors. A \textbf{lower} score signifies greater identity diversity.
\begin{equation}
\label{eq:ids}
\text{IDS} = \frac{2}{N(N-1)} \sum_{i=1}^{N} \sum_{j=i+1}^{N} \frac{v_i \cdot v_j}{\|v_i\| \|v_j\|}
\end{equation}

\noindent\textbf{Artistic Style Coverage (ASC):} 
Inspired by the Image Retrieval Score (IRS)\cite{irs}, this metric quantifies the retrievable Style diversity of a generative model relative to the ground-truth data's diversity. We use the CSD\cite{csd} feature extractor ($\mathcal{F}$) and define three datasets: a ground-truth (GT) training set $\mathcal{X}_{train}$ (with $N_{train}$ images from WikiArt\cite{wikiart}), a GT reference set $\mathcal{X}_{test}$ (with $N_{sample}=N_{test}$ from WikiArt\cite{wikiart}), and the generated synthetic set $\mathcal{X}_{synth}$ (with $N_{sample}=N_{synth}$ from \bench).The score is computed in three steps:

\begin{itemize}
    \item \textbf{Retrieval:} For a query set $\mathcal{X}_{q}$ (either $\mathcal{X}_{test}$ or $\mathcal{X}_{synth}$), we first find $\mathcal{X}_{learned}$, the set of unique training images that are the nearest neighbor (in CSD feature space $\mathcal{F}$) to at least one image in $\mathcal{X}_{q}$, as defined in Eq.~\eqref{eq:asc_xlearned}. We then get the count $N_{learned} = |\mathcal{X}_{learned}(\mathcal{X}_{q})|$.
    \begin{equation}
    \label{eq:asc_xlearned}
    \begin{split}
        \mathcal{X}_{learned}(\mathcal{X}_{q}) = \{ x \in \mathcal{X}_{train} \mid \exists g \in \mathcal{X}_{q} \text{ s.t. } \\
        x = \text{arg min}_{x' \in \mathcal{X}_{train}} d(\mathcal{F}(g), \mathcal{F}(x')) \}
    \end{split}
    \end{equation}

    \item \textbf{Estimation:} Using $N_{learned}$, $N_{sample}$, and $N_{train}$, we compute the maximum likelihood estimate of the total "learnable" images $s^*$ (Eq.~\eqref{eq:asc_s_star}), and the corresponding "infinite" retrieval score, $IRS_{\infty}$ (Eq.~\eqref{eq:asc_irs_inf}). This estimates the fraction of $\mathcal{X}_{train}$ that would be retrieved given infinite query samples.
    \begin{equation}
    \label{eq:asc_s_star}
    s^*(\mathcal{X}_{q}) = \text{arg max}_{s} P(N_{learned}, N_{sample}, s)
    \end{equation}
    \begin{equation}
    \label{eq:asc_irs_inf}
    IRS_{\infty}(\mathcal{X}_{q}) = s^*(\mathcal{X}_{q}) / N_{train}
    \end{equation}
    
    \item  \textbf{Adjustment:} To correct for the feature extractor's inherent "measurement gap", the final ASC score is the \textbf{Adjusted IRS} ($IRS_{\infty, a}$), shown in Eq.~\eqref{eq:asc}. This is the ratio of the synthetic set's estimated diversity to the real reference set's estimated diversity. A \textbf{higher} score is better.
    \begin{equation}
    \label{eq:asc}
    \text{ASC} = \frac{IRS_{\infty}(\mathcal{X}_{synth})}{IRS_{\infty}(\mathcal{X}_{test})}
    \end{equation}
    
\end{itemize}

\begin{table*}[t]
\centering
\caption{\textbf{Prompts Used for The Qualitative Examples in Figure 1 of The Main Paper Grouped by Their Corresponding Dimension.} The numbering (1-16) corresponds to the images in Figure 1, read from left-to-right, top-to-bottom within each dimensional category.}
\label{tab:fig1_prompts}
\small 
\begin{tabularx}{\textwidth}{c c X} 
\toprule
\textbf{No.} & \textbf{Dimension} & \textbf{Prompt} \\
\midrule

1 & \multirow{4}{*}{Face} & A high-quality portrait photo of a young Middle Eastern woman who is attractive with arched eyebrows \\
2 & & A high-quality portrait photo of a middle-aged Middle Eastern woman with a receding hairline who is attractive \\
3 & & A high-quality portrait photo of a middle-aged White woman with a big nose \\
4 & & A high-quality portrait photo of a young Middle Eastern woman with an oval face \\
\midrule 

5 & \multirow{4}{*}{Style} & An artwork of corgi pizza, in the Baroque style. \\
6 & & Imagine a panda bear playing ping pong using a blue paddle against an ostrich using a red paddle. Now, picture it in the style of Fauvism. \\
7 & & An image of a giant cobra snake made from salad, with strong Action painting influences. \\
8 & & A masterpiece of Pointillism, showing a hot air balloon \\
\midrule 

9-12 & Layout & A clear, top-down view of two tennis rackets arranged on a large white table. \\
\midrule 

13 & \multirow{4}{*}{Tonal} & a woman with sunglasses and red hair, monochrome, black and white \\
14 & & a tiger in a forest, desaturated, muted colors \\
15 & & An image of an ornate treasure chest with a broad sword propped up against it, glowing in a dark cave, rendered with natural colors properties \\
16 & & Photograph of three red lego blocks, captured with neon colors, fluorescent \\
\bottomrule
\end{tabularx}
\end{table*}

\begin{figure*}[!t]
  \centering
    \includegraphics[width=1.0\textwidth]{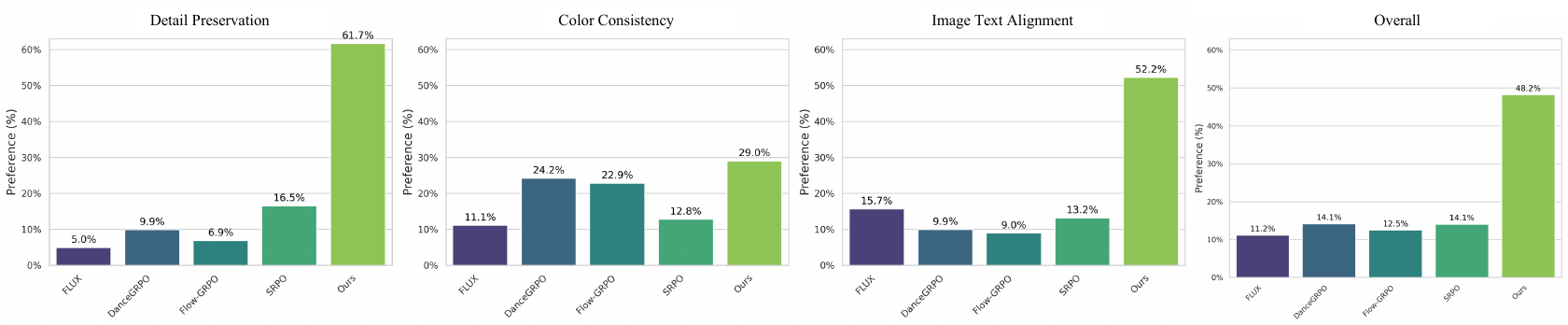}
    
    \caption{\textbf{Human Preference Evaluation on HPDv2.} We conducted a user study comparing \ourshort against the base model FLUX and state-of-the-art RL alignment methods (DanceGRPO, Flow-GRPO, SRPO). The evaluation spans four distinct dimensions: Detail Preservation, Color Consistency, Image-Text Alignment, and Overall Preference. \ourshort achieves a dominant lead in Detail Preservation ($61.7\%$) and Image-Text Alignment ($52.2\%$), significantly outperforming baselines that suffer from mode collapse artifacts. Ultimately, our method secures the highest Overall Preference rate of $48.2\%$.}
    \vspace{-15pt}
    \label{fig: userstudy}
\end{figure*}


\noindent\textbf{Spatial Dispersion Index (SDI):} This metric evaluates the diversity of object layouts across multiple images generated from the \emph{same} text prompt, effectively measuring the model's ability to produce spatially varied results. For $M$ images per prompt, we first use Grounding DINO\cite{groundingdino} to detect the bounding boxes $L_i=\{b_j\}$ of the target objects in each image. The Similarity $\text{Sim}_{\text{Layout}}$ (Eq.~\ref{eq:sim_layout}) is calculated by finding the optimal bipartite matching of the detected bounding boxes via the Hungarian algorithm on the IoU matrix, normalized by the maximum number of objects. We then compute the average pairwise \textbf{Layout Similarity ($\overline{\text{Sim}}_{\text{Layout}}$)} between all pairs of images, as defined in Eq.~\ref{eq:avg_sim_layout}. Finally, the \textbf{SDI} is defined as one minus the average Layout Similarity across all $M$ images, averaged over all prompts $P$ (Eq.~\ref{eq:sdi}). A \textbf{higher} score signifies greater Layout diversity.

\begin{figure*}[!t]
  \centering
    \includegraphics[width=1.0\textwidth]{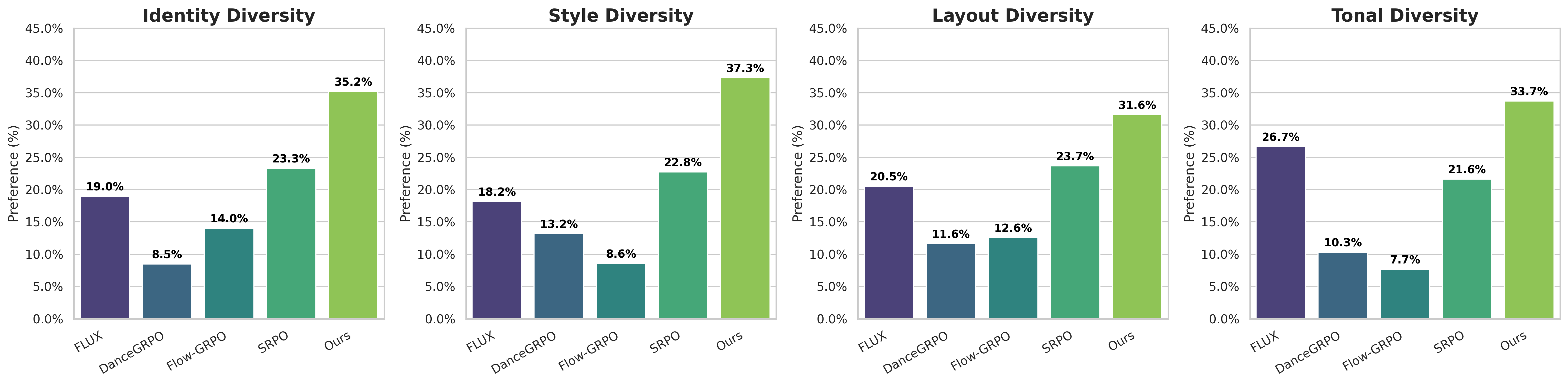}
    
    \caption{\textbf{Human Preference on Diversity (DivGenBench).} We evaluated user preferences across four key diversity dimensions: Identity, Style, Layout, and Tonal. The results reveal a severe PMC in existing RL baselines (DanceGRPO, Flow-GRPO), which often score lower than the Base Model (FLUX), particularly in Tonal and Style diversity. In contrast, \ourshort consistently achieves the highest preference rates (e.g., $37.3\%$ in Style and $35.2\%$ in Identity), demonstrating its ability to break the trade-off between human preference and generative diversity.}
    \vspace{-15pt}
    \label{fig: userstudy_diversity}
\end{figure*}

\begin{equation}
\label{eq:sim_layout}
\begin{aligned}
\text{Sim}_{\text{Layout}}(L_i, L_p) = & \frac{1}{\max(|L_i|, |L_p|)} \\
& \times \sum_{(j, l) \in \mathcal{P}} \text{IoU}(b_j \in L_i, b_l \in L_p)
\end{aligned}
\end{equation}
where $\mathcal{P}$ is the set of optimal matching pairs found via the Hungarian algorithm.

\begin{equation}
\label{eq:avg_sim_layout}
\overline{\text{Sim}}_{\text{Layout}} = \frac{2}{M(M-1)} \sum_{i=1}^{M} \sum_{p=i+1}^{M} \text{Sim}_{\text{Layout}}(L_i, L_p)
\end{equation}

\begin{equation}
\label{eq:sdi}
\text{SDI} = \frac{1}{P} \sum_{r=1}^{P} \left( 1 - \overline{\text{Sim}}_{\text{Layout}}^{(r)} \right)
\end{equation}

\noindent\textbf{Photographic Variance Score (PVS):} This metric, inspired by APG\cite{apg}, quantifies the spread of generated tonal values. For a set of $N$ images $G = \{g_i\}_{i=1}^N$, we first extract a scalar value for each perceptual dimension. For each image $g_i$, \textbf{Saturation} ($s_i$) is the mean of the S-channel (from an RGB-to-HSV conversion), \textbf{Brightness} ($v_i$) is the mean of the V-channel, and \textbf{Contrast} ($c_i$) is the standard deviation of the grayscale-converted image. We then form three value sets $\mathbf{s} = \{s_i\}_{i=1}^N$, $\mathbf{v} = \{v_i\}_{i=1}^N$, and $\mathbf{c} = \{c_i\}_{i=1}^N$. PVS is the sum of the standard deviations of these three sets. A \textbf{higher} score indicates greater tonal control.
\begin{equation}
\label{eq:pvs}
\text{PVS} = \text{std}(\mathbf{s}) + \text{std}(\mathbf{v}) + \text{std}(\mathbf{c})
\end{equation}

\section{Extended Experiments}
\label{sec:extended_results}

\subsection{User Study on HPDv2}
\label{sec:user_study_results}
To validate the effectiveness of \ourshort in alignment with human preferences, we conducted a comprehensive user study following \cite{dancegrpo, chen2025s2guidancestochasticselfguidance}. We compared our method against the base model (FLUX) and three competitive RL-based baselines: DanceGRPO, Flow-GRPO, and SRPO.

\subsubsection{Experimental Setup}
We randomly selected \texttt{100} prompts from \textit{HPDv2}. For each prompt, we generated images using all five methods with the same random seed. We recruited \texttt{20} evaluators who were presented with the generated images in a randomized, blind manner. The evaluators were asked to select the best image based on the following criteria:
\begin{itemize}
    \item \textbf{Detail Preservation:} The clarity, sharpness, and richness of details in the generated image.
    \item \textbf{Color Consistency:} The naturalness, harmony, and realism of the colors.
    \item \textbf{Image-Text Alignment:} How well the generated image accurately reflects the content and intent of the text prompt.
    \item \textbf{Overall:} Considering all the above factors, which image do you prefer?
\end{itemize}

\subsubsection{Analysis of Results}
The results of the user study are presented in Figure~\ref{fig: userstudy}. The findings demonstrate a clear and consistent preference for our proposed method, \ourshort, across all evaluated metrics. Specifically, in the \textit{Detail Preservation} category, \ourshort was preferred in 61.7\% of cases, significantly outperforming the runner-up, SRPO (16.5\%). A similar dominant trend is observed for \textit{Image-Text Alignment}, where \ourshort achieved a 52.2\% preference rate. Furthermore, for \textit{Color Consistency}, our method was chosen 29.0\% of the time, again marking a lead over all baselines.

Aggregating the votes, the \textit{Overall} preference for \ourshort stands at 48.2\%, confirming its comprehensive superiority. This strong performance in human evaluations validates that \ourshort not only improves alignment from a theoretical standpoint but also translates to tangible and perceptually superior generation quality that is easily recognized by human users.

\subsection{User Study on \bench}
\label{sec:user_study_divgen}

While the HPDv2 study confirms our method's alignment quality, it does not explicitly measure the severity of PMC. To quantitatively assess whether models sacrifice diversity for higher scores, we conducted a second user study using our proposed \textit{DivGenBench}.

\subsubsection{Experimental Setup}
We sampled \texttt{20} distinct templates from each of the four dimensions in DivGenBench: \textit{Identity}, \textit{Style}, \textit{Layout}, and \textit{Tonal}, totaling \texttt{80} evaluation sets. For each set, we generated images using varied prompts designed to probe the model's generative boundaries (e.g., requesting specific "Low-key" lighting or distinct "Cubism" styles). \texttt{20} evaluators were asked to identify which model best reflected the requested diversity and avoided generating repetitive or homogeneous outputs.

\subsubsection{Analysis of Results}
The diversity preference results are presented in Figure~\ref{fig: userstudy_diversity}. Two critical observations emerge from the data:

\paragraph{Evidence of Preference Mode Collapse.}
The results provide strong empirical evidence of PMC in existing RL methods. In several dimensions, the baseline RL models (DanceGRPO and Flow-GRPO) perform significantly worse than the unaligned Base Model (FLUX). For instance, in \textit{Tonal Diversity}, Flow-GRPO drops to a mere 7.7\% preference rate, and DanceGRPO to 10.3\%, compared to FLUX's 26.7\%. Similarly, in \textit{Style Diversity}, Flow-GRPO (8.6\%) lags behind FLUX (18.2\%). This confirms that naively optimizing for reward models drives the generator into a narrow mode (e.g., always generating over-exposed or realistic styles), actively destroying the inherent diversity of the base model.

\paragraph{Superior Diversity Preservation.}
In contrast, \ourshort effectively mitigates this collapse. Our method achieves the highest preference scores across all four dimensions, surpassing both the collapsed baselines and the Base Model.
\begin{itemize}
    \item \textbf{Identity \& Style:} We achieve dominant preference rates of 35.2\% for \textit{Identity} and 37.3\% for \textit{Style}, significantly outperforming the runner-up SRPO (about 23\%). This indicates our method can generate diverse faces and artistic styles without reverting to a mean template.
    \item \textbf{Layout \& Tonal:} Crucially, in the dimensions most susceptible to collapse, our method maintains robustness. In \textit{Tonal Diversity}, where baselines fail, \ourshort leads with 33.7\%, demonstrating successful disentanglement of "quality" from "lighting bias."
\end{itemize}
These results, combined with the HPDv2 findings, validate that \ourshort can simultaneously improve human preference alignment while preserving and even enhancing generative diversity.

\begin{table*}[!t]
\centering
\caption{\textbf{Comprehensive Quantitative Evaluation of Metrics for Human Preference Alignment and Semantic Consistency}. We compare FLUX, DanceGRPO, and DanceGRPO incorporated with our learned $\boldsymbol{b}_v$. All RL-based methods utilize HPS-v2.1 as the reward model. \textbf{Ranking is performed between the RL-based methods.} The best score is shown in \textbf{bold}.}
\label{tab:quantitative_evaluation_combined_abaltion}
\resizebox{\textwidth}{!}{ 
\begin{tabular}{l *{8}{c}} 
\toprule
\multicolumn{1}{l|}{\multirow{3}{*}{\textbf{Method}}} & \multicolumn{5}{c|}{\textbf{Human Preference Alignment}} & \multicolumn{3}{c}{\textbf{Semantic Consistency \& Accuracy}} \\
\cmidrule(lr){2-6} \cmidrule(lr){7-9}
\multicolumn{1}{l|}{} & {\makecell{Aesthetic $\uparrow$}} & {\makecell{ImageReward $\uparrow$}} & {\makecell{Pick Score $\uparrow$}} & {\makecell{Q-Align $\uparrow$}} & \multicolumn{1}{c|}{\makecell{HPS-v2.1 $\uparrow$}} & {\makecell{CLIP $\uparrow$}} & {\makecell{DeQA $\uparrow$}} & \multicolumn{1}{c}{\makecell{GenEval $\uparrow$}} \\
\midrule
\multicolumn{1}{l|}{FLUX} & 6.417 & 1.670 & 0.240 & 4.922 & 0.310 & 0.315 & 4.456 & 0.663 \\
\midrule
\multicolumn{1}{l|}{DanceGRPO} & 6.068 & 1.664 & 0.241 & 4.930 & \textbf{0.361} & 0.293 & 4.400 & 0.522 \\
\multicolumn{1}{l|}{DanceGRPO + Our learned $\boldsymbol{b}_v$} & \textbf{6.353} & \textbf{1.677} & \textbf{0.242} & \textbf{4.947} & 0.319 & \textbf{0.311} & \textbf{4.496} & \textbf{0.641} \\
\bottomrule
\end{tabular}
} 
\vspace{-10pt}
\end{table*}

\subsection{Generalizability Study with DanceGRPO}
\label{subsec:generalizability_dancegrpo}

To further evaluate the intrinsic value and generalizability of our learned corrective signal $\mathbf{b}_{v}$, we conducted an extension experiment by applying it as a plug-and-play component to an external Reinforcement Learning framework. Specifically, we selected DanceGRPO~\cite{dancegrpo}, a representative method that, while effective in improving alignment, is susceptible to PMC. The objective of this experiment is to verify whether the directionality captured by $\mathbf{b}_{v}$ acts as a universal corrective signal for the reward model and can effectively mitigate mode collapse in other algorithms.

\subsubsection{Experimental Setup}
We maintain the original training logic and hyperparameters of DanceGRPO. The only modification lies in the reward calculation mechanism. Let $\mathbf{b}_v^*$ denote the optimal parameter learned and frozen from Stage 1 of our method. We substitute the naive reward of DanceGRPO with our proposed guided reward $R_{\text{guided}}$. Formally, during the DanceGRPO training process, for every generated sample $(\vx_0, c)$, the reward is computed as:
\begin{equation}
    R_{\text{guided}}(\vx_0, c; \mathbf{b}_v^*) = \score(\Phi_\text{img}(\vx_0), \tilde{\boldsymbol{e}}_{\text{text}})
\end{equation}
where $\tilde{\boldsymbol{e}}_{\text{text}}$ is the rectified text embedding constructed using $\mathbf{b}_v^*$ via Eq.~\eqref{eq:guided_embedding} (as defined in the main paper), with the guidance scale $\omega$ kept consistent.

\subsubsection{Analysis of Results}
The quantitative comparisons in Tab.~\ref{tab:quantitative_evaluation_combined_abaltion} and Tab.~\ref{tab:quantitative_evaluation_diversity_single_col_abaltion} demonstrate that integrating our corrective signal significantly enhances the robustness of DanceGRPO.
In terms of alignment, while vanilla DanceGRPO achieves the highest HPS-v2.1 score, it suffers from regression in generalized metrics. In contrast, applying our learned $\mathbf{b}_v$ effectively mitigates reward overfitting: it achieves a \textbf{4.7\%} improvement in Aesthetic Score and restores semantic consistency with a \textbf{6.1\%} gain in CLIP score, suggesting a shift towards true human preference.
Crucially, for diversity, our method effectively resolves the mode collapse observed in the baseline. By filtering out the low-diversity manifold, our corrective signal reduces the IDS score by \textbf{20.1\%} and expands the ASC score by a remarkable \textbf{57.7\%}, surpassing both the baseline and the pre-trained FLUX. 
These results confirm that $\mathbf{b}_v$ forces the external optimizer to explore a broader solution space without requiring complex re-tuning of the training configuration.

\begin{table}[htb]
\centering
\caption{\textbf{Quantitative Evaluation of Generative Diversity on \bench}. We compare FLUX, DanceGRPO, and DanceGRPO enhanced with our learned $\boldsymbol{b}_v$. All RL-based methods utilize HPS-v2.1 as the reward model. We report Identity Divergence Score (IDS), Artistic Style Coverage (ASC), Spatial Dispersion Index (SDI), and Photographic Variance Score (PVS). \textbf{Ranking is performed between the RL-based methods.} The best score is shown in \textbf{bold}.
}
\label{tab:quantitative_evaluation_diversity_single_col_abaltion}
\resizebox{\linewidth}{!}{
\begin{tabular}{l *{4}{c}}
\toprule
\multicolumn{1}{l|}{\textbf{Method}} & {\makecell{IDS $\downarrow$}} & {\makecell{ASC $\uparrow$}} & {\makecell{SDI $\uparrow$}} & {\makecell{PVS $\uparrow$}} \\
\midrule
\multicolumn{1}{l|}{FLUX} & 0.280 & 0.179 & 0.563 & 0.408 \\
\midrule
\multicolumn{1}{l|}{DanceGRPO} & 0.348 & 0.130 & 0.488 & 0.259 \\
\multicolumn{1}{l|}{DanceGRPO + Our learned $\boldsymbol{b}_v$} & \textbf{0.278} & \textbf{0.205} & \textbf{0.604} & \textbf{0.437} \\
\bottomrule
\end{tabular}
}
\vspace{-10pt}
\end{table}


\section{Visualization}
\label{sec:visualization}

\subsection{Prompts for Figure 1}
We present the example prompts used to generate the qualitative comparisons in Figure 1 of the main paper. The prompts, grouped by their corresponding dimension, are detailed in Tab.~\ref{tab:fig1_prompts}.

\subsection{Results on HPDv2}
\label{sec:qualitative_hpdv2}

\begin{figure*}[!t]
  \centering
    \includegraphics[width=1.0\textwidth, height=1.0\textheight, keepaspectratio]{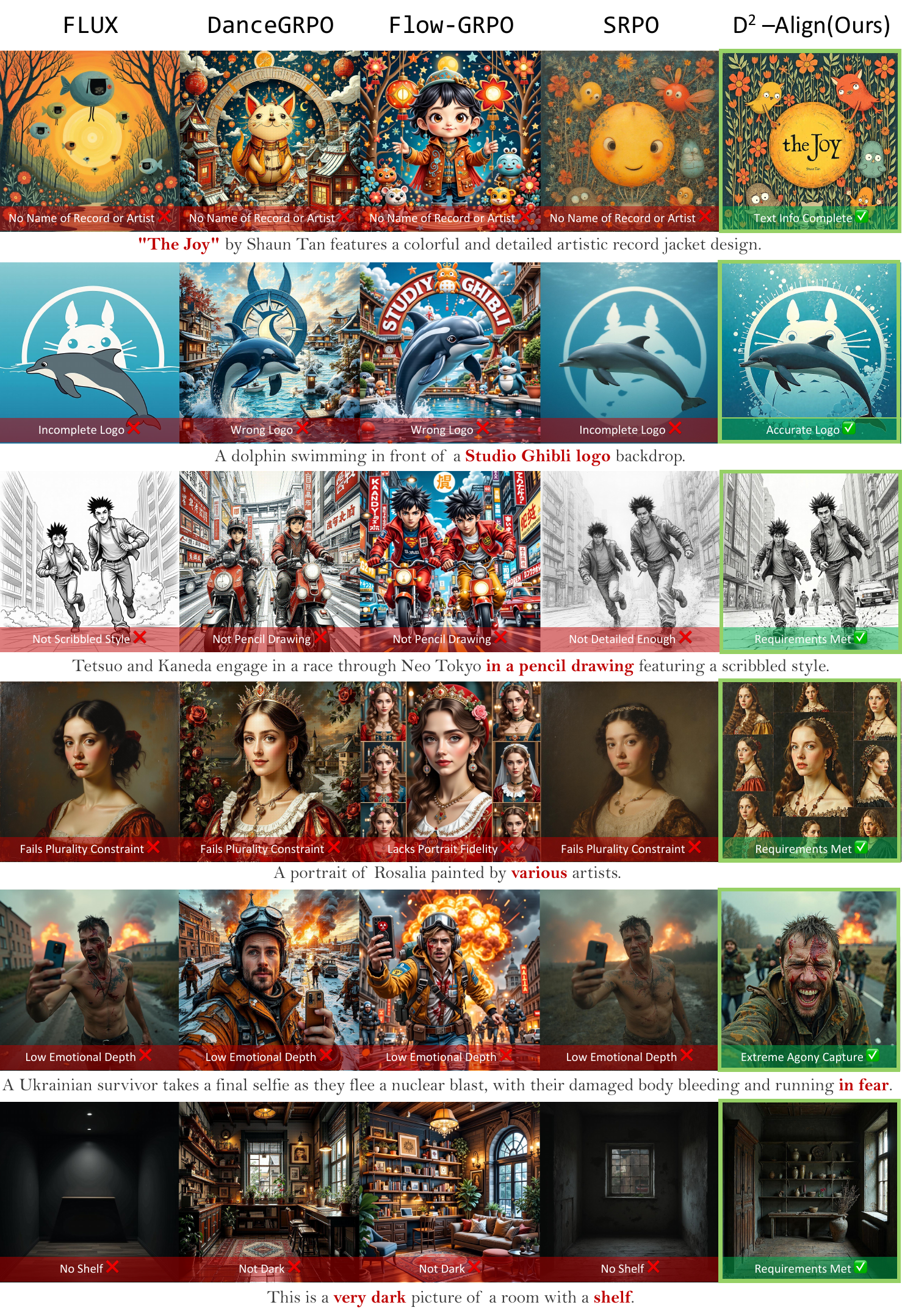}
    \caption{Qualitative comparison of \ourshort against SOTAs on the HPDv2 benchmark. All RL-based methods are trained using HPS-v2.1 as the reward model.}
    \vspace{-10pt}
    \label{fig:hpdv2_hps_only}
\end{figure*}

\begin{figure*}[!t]
  \centering
    \includegraphics[width=1.0\textwidth, height=1.0\textheight, keepaspectratio]{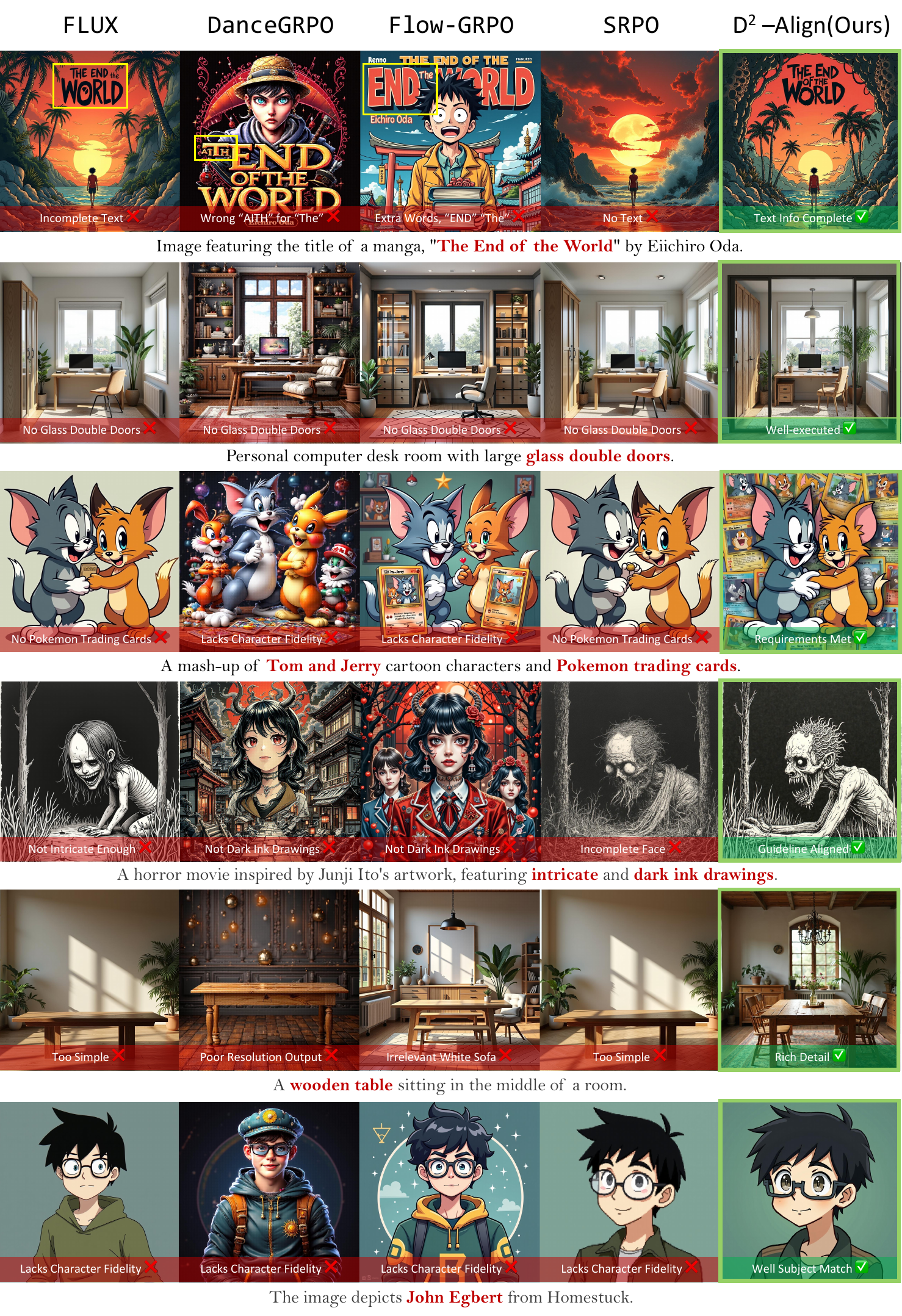}
    \caption{Qualitative comparison of \ourshort against SOTAs on the HPDv2 benchmark. All RL-based methods are trained using HPS-v2.1 and CLIP as the reward models.}
    \vspace{-10pt}
    \label{fig:hpdv2_hps_clip}
\end{figure*}

In this section, we present qualitative comparisons on the HPDv2\cite{hps} benchmark to visually evaluate the performance of our method against the baseline approach and advanced RL-based methods. Fig.\ref{fig:hpdv2_hps_only} illustrates the visual outputs where all competing RL-based methods were trained using only HPS-v2.1\cite{hps} as the reward model. Fig.\ref{fig:hpdv2_hps_clip} shows the results for the same set of methods, but where the models were trained using a combined reward signal from both HPS-v2.1\cite{hps} and CLIP\cite{hessel2021clipscore}, inspired by DanceGRPO\cite{dancegrpo}.

\subsection{Results on \bench}
\label{sec:qualitative_DivGenBench}

We provide a comprehensive visual evaluation of our method on \bench. Observing a critical trade-off in existing work, we strategically focus our comparative visualization on the two state-of-the-art methods that achieved the highest performance on the HPDv2 benchmark yet simultaneously recorded the lowest diversity scores on \bench. 

\begin{figure*}[!t]
  \centering
  \includegraphics[width=1.0\textwidth, height=1.0\textheight, keepaspectratio]{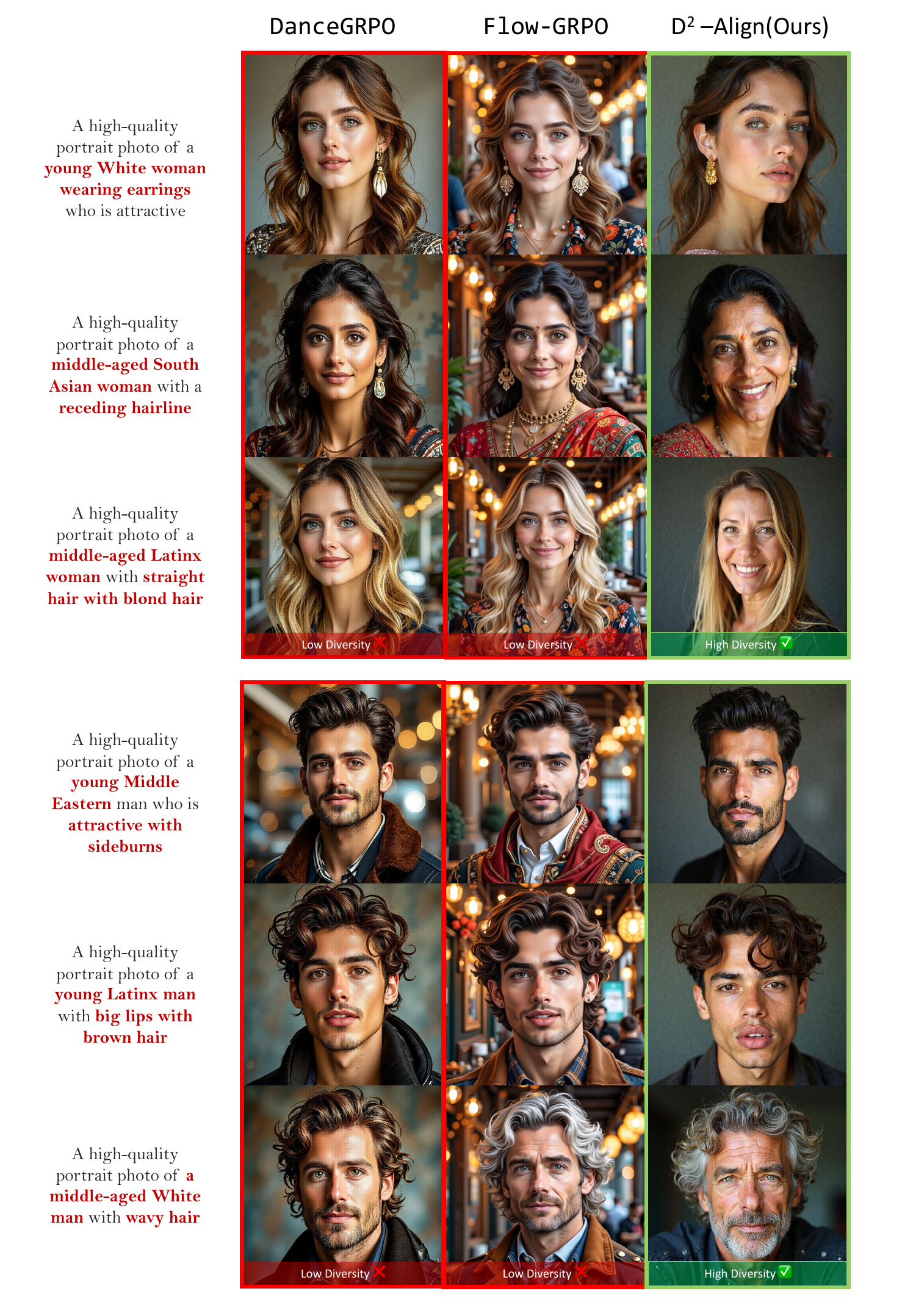} 
  \caption{Qualitative comparison on the ID dimension of \bench. Our method generates diverse identities adhering to required demographic features.}
  \vspace{-10pt}
  \label{fig:bench_id}
\end{figure*}

\begin{figure*}[!t]
  \centering
  \includegraphics[width=1.0\textwidth, height=1.0\textheight, keepaspectratio]{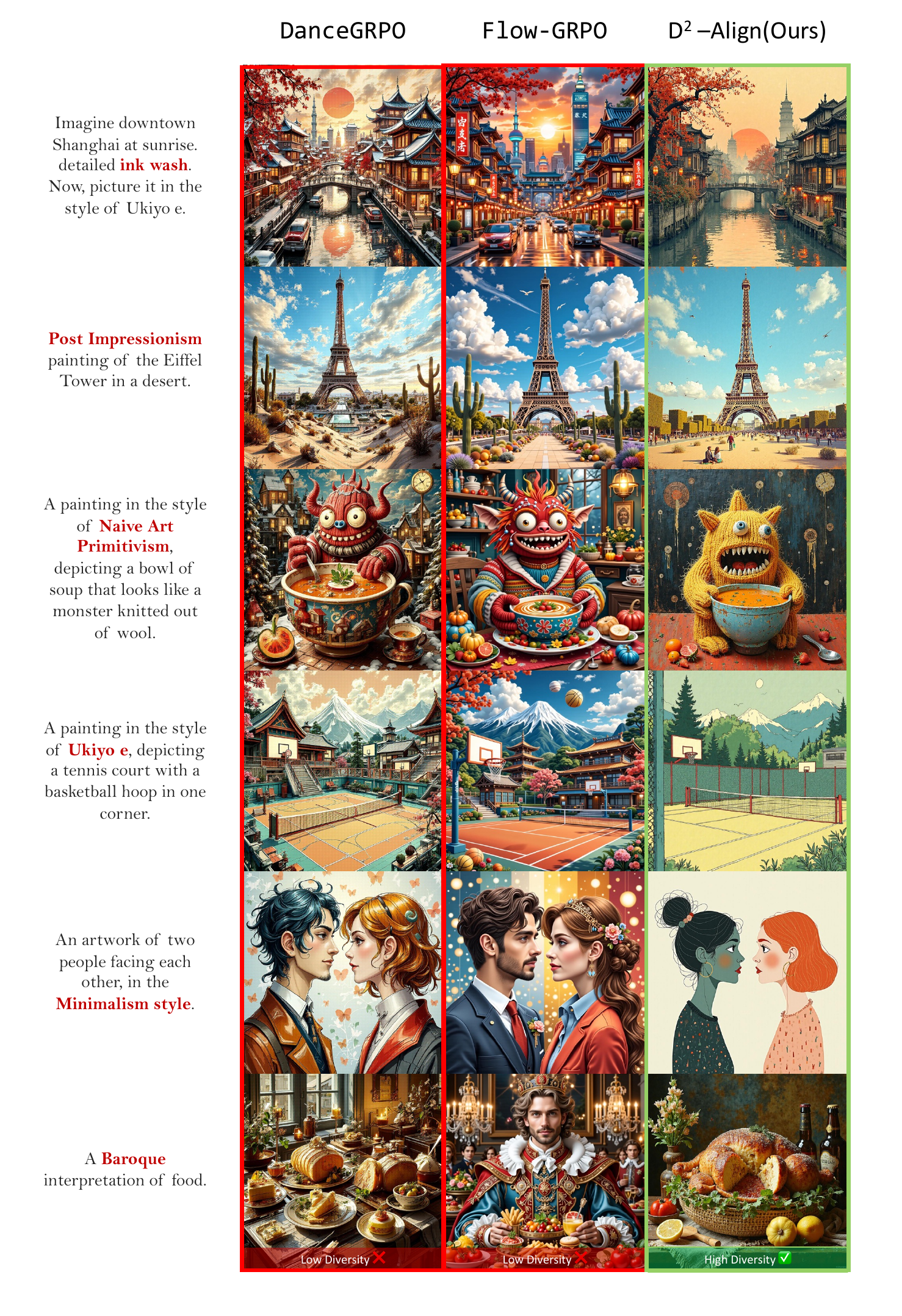} 
  \caption{Qualitative comparison on the Style dimension of \bench. Our method faithfully renders diverse artistic styles specified in the prompts.}
  \vspace{-10pt}
  \label{fig:bench_style}
\end{figure*}

\begin{figure*}[!t]
  \centering
  \includegraphics[width=1.0\textwidth, height=1.0\textheight, keepaspectratio]{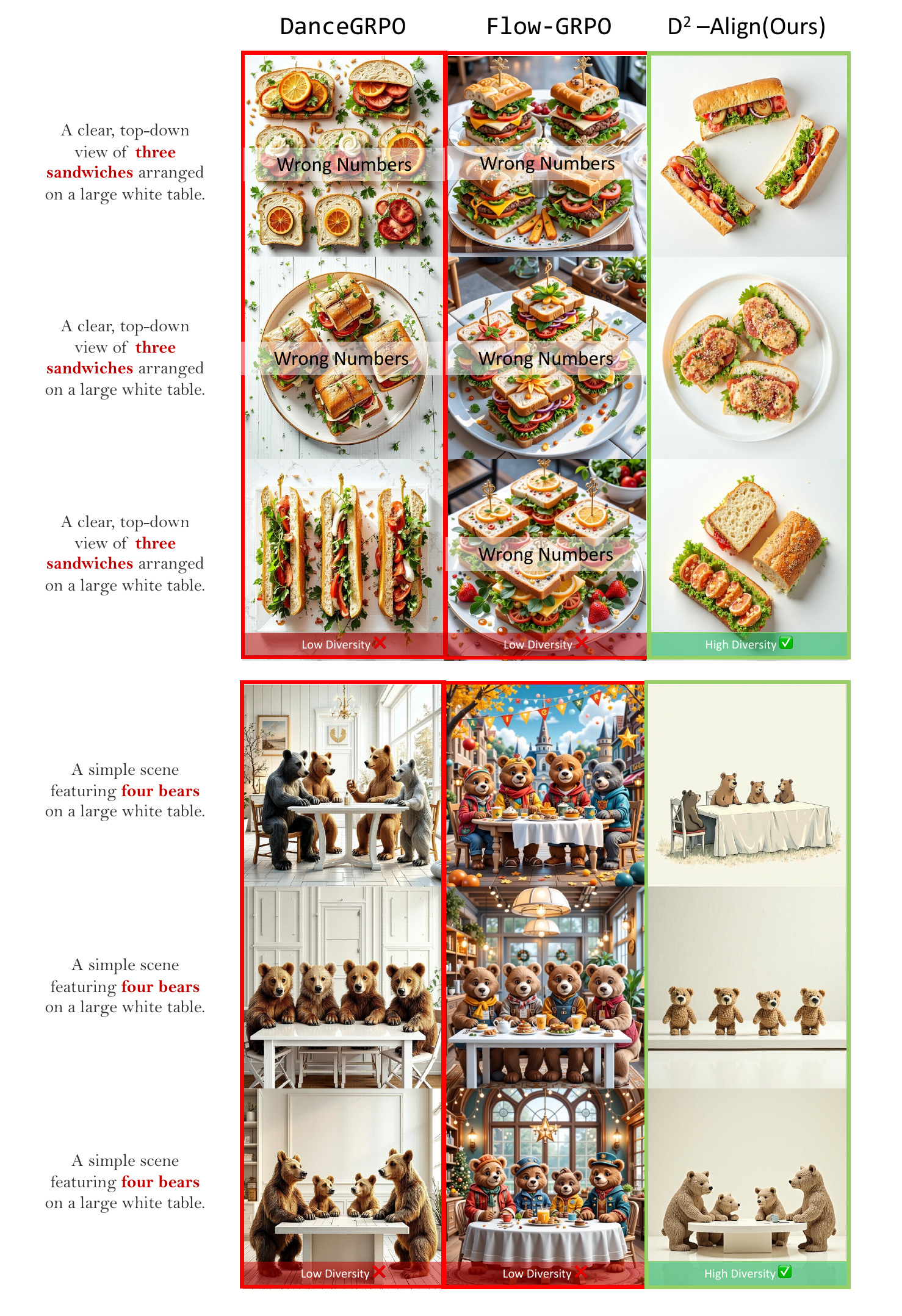} 
  \caption{Qualitative comparison on the Layout dimension of \bench. Our method not only achieves precise adherence to object counts but also generates diverse and novel spatial arrangements.}
  \vspace{-10pt}
  \label{fig:bench_layout}
\end{figure*}

\begin{figure*}[!t]
  \centering
  \includegraphics[width=1.0\textwidth, height=1.0\textheight, keepaspectratio]{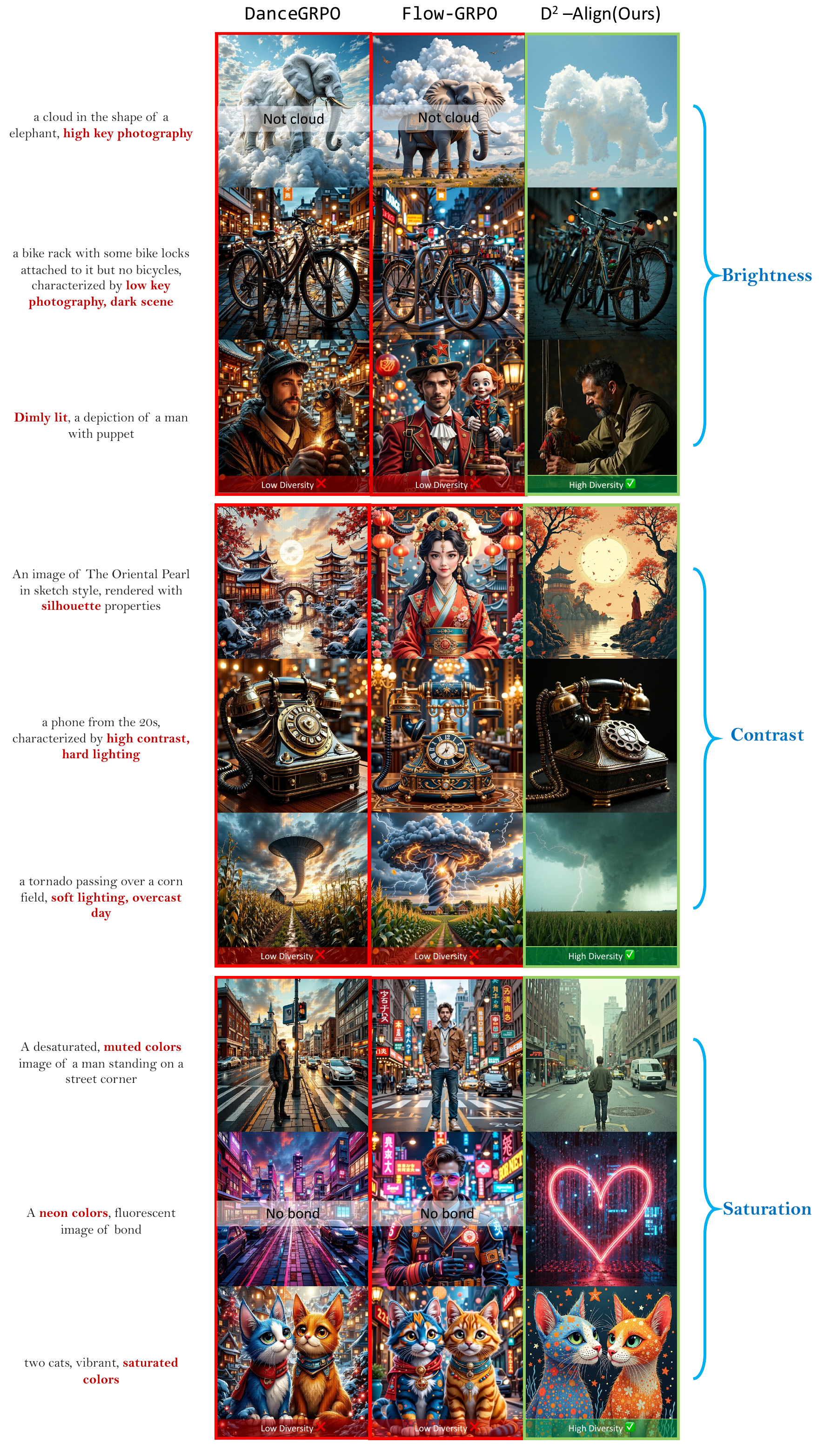} 
  \caption{Qualitative comparison on the Tonal dimension of \bench. Our method generates a diverse spectrum of brightness, contrast, and saturation levels while maintaining high image fidelity.}
  \vspace{-10pt}
  \label{fig:bench_tonal}
\end{figure*}

As shown in Fig.\ref{fig:bench_id}, our method generates distinct identities, avoiding the mode collapse seen in baselines. It further demonstrates a broad range of artistic styles without defaulting to generic aesthetics (Fig.\ref{fig:bench_style}), produces diverse spatial layouts (Fig.\ref{fig:bench_layout}), and maintains a wide tonal spectrum in brightness and contrast (Fig.\ref{fig:bench_tonal}), contrasting with the monotonic distributions of competing methods.


\end{document}